\pdfoutput=1

\documentclass[11pt]{article}

\usepackage[final]{acl}

\usepackage{times}
\usepackage{latexsym}

\usepackage{listings}
\usepackage{amsmath} 
\usepackage{subcaption}
\usepackage[T1]{fontenc}

\usepackage[utf8]{inputenc}

\usepackage{microtype}

\usepackage{inconsolata}

\usepackage{graphicx}
\usepackage{tcolorbox}
\usepackage{enumitem}
\usepackage{amssymb}
\usepackage{booktabs} 
\usepackage{float}
\usepackage{multirow} 

\title{FiVL: A Framework for Improved Vision-Language Alignment through the Lens of Training, Evaluation and Explainability}

\author{
 \textbf{Estelle Aflalo\textsuperscript{*,1}},
 \textbf{Gabriela Ben Melech Stan \textsuperscript{*,1}},
 \textbf{Tiep Le\textsuperscript{1}},
 \textbf{Man Luo \textsuperscript{1}},
\\
 \textbf{Shachar Rosenman \textsuperscript{1}},
 \textbf{Sayak Paul \textsuperscript{2}},
 \textbf{Shao-Yen Tseng\textsuperscript{1}},
 \textbf{Vasudev Lal \textsuperscript{1}},
\\
\\
 \textsuperscript{*} Equal contribution,
 \textsuperscript{1}Intel Labs,
 \textsuperscript{2}Hugging Face,
\\
 \small{
   \textbf{Correspondence:} \href{mailto:email@domain}{estelle.aflalo@intel.com}
 }
}

\begin{document}
\maketitle
\begin{abstract}

Large Vision Language Models (LVLMs) have achieved significant progress in integrating visual and textual inputs for multimodal reasoning. 
However, a recurring challenge is ensuring these models utilize visual information as effectively as linguistic content when both modalities are necessary to formulate an accurate answer. We hypothesize that hallucinations arise due to the lack of effective visual grounding in current LVLMs.
Furthermore, current vision-language benchmarks are not specifically measuring the degree to which the answer require the visual input. This limitation makes it challenging to confirm that the image is truly necessary, particularly in tasks like visual question answering. In this work, we introduce FiVL, a novel method for constructing datasets designed to train LVLMs for enhanced visual grounding and also evaluate their effectiveness in achieving it. 
We demonstrate the value of our datasets through three approaches. First, we introduce a novel training task based on our augmented training dataset, resulting in better performance than the baseline. Second, we present benchmarks to assess the model's ability to use image as substantive evidence, rather than relying solely on linguistic priors. Finally, we identify attention heads with the strongest vision-language alignment, enabling explainability on visual-driven hallucinations.
The dataset and code will be publicly available.  

\end{abstract}

\section{Introduction}
\label{sec:intro}
Recent advancements in large language models have led to the integration of non-linguistic information through multimodal perception and generation, culminating in the development of Large Vision Language Models (LVLM). These models effectively bridge visual comprehension and linguistic reasoning, offering a unified approach to multimodal understanding and instruction-following \cite{liu2024visual, peng2023kosmos, wang2024qwen2vlenhancingvisionlanguagemodels, openai2024gpt4ocard}. 
However, despite their apparent adeptness in visual perception, LVLMs still face the challenge of ``hallucination'' — where the model generates semantically plausible yet factually incorrect information that is inconsistent with the image. 
This issue possibly arises from the imbalance of visual data compared to text data during training, which limits the model's ability to overcome preconceptions inherited from the underlying LLM \cite{liu2024survey}.
A common approach to mitigate this issue has been to introduce a visual grounding mechanism to the model \cite{chen2023shikra, peng2023kosmos, wang2024qwen2vlenhancingvisionlanguagemodels,rasheed2024glamm, Zhang_2024_CVPR, zhang2024llava}. 

Visual grounding aims to achieve more precise alignment between visual attention and semantic concepts within the model. 
A common method for grounding involves using bounding boxes, represented as a sequence of numerical numbers, to specify a particular region of an image. 
This enables the user to query specific parts of the image and the model to reference image locations within its generated response \cite{chen2023shikra, peng2023kosmos, wang2024qwen2vlenhancingvisionlanguagemodels}.
Bounding boxes, however, are coarse coordinates and unable to highlight objects or abstract concepts in finer detail. 
Recent work have addressed these concerns by applying pixel-level grounding through the use of segmentation masks instead \cite{rasheed2024glamm, Zhang_2024_CVPR, zhang2024llava}. 

Training models with pixel-level grounding requires datasets that provide fine-grained visual alignment between images and text.  
However, such datasets are scarce, and prior work have often constructed custom datasets alongside model development \cite{zhang2024llava, rasheed2024glamm, ma2024groma, Zhang_2024_CVPR}. 
Additionally, prior works use grounding datasets only for training and overlook the importance of alignment datasets for evaluation.
To address these challenges, we introduce FiVL, a novel Framework towards Improved Vision-Language Alignment, for constructing datasets with visual-concept alignment. We demonstrate the usefulness of these datasets through a novel training approach as well as a method that evaluates and interprets the visual-language alignment capability in LVLMs. 


Our main contributions are as follows: 
\begin{itemize}[noitemsep,leftmargin=*]
\item We introduce FiVL, a framework designed to augment multimodal datasets with vision-alignment capabilities. Through comprehensive human and automated evaluations of the datasets produced by FiVL, we demonstrate their reliability. 

\item Using the FiVL training dataset, we introduce a novel training task that jointly trains text and vision tokens. Leveraging this task, we fine-tuned an LVLM model that outperforms the baseline across several downstream tasks.

\item Through FiVL evaluation datasets, we use a perturbation-based approach to assess the vision-alignment capability of LVLMs and introduce visual reliance score. This score shows a strong correlation with overall model performance, going beyond a specific subset of benchmarks.
\item We leverage our framework in order to gain more insights into the internal mechanisms of LVLMs by identifying attention heads with the strongest vision-language alignment capabilities, as demonstrated in \cite{aflalo2022vlinterpretinteractivevisualizationtool}. This approach enables the exploration of vision-based hallucinations.
\end{itemize}
\section{Related Work}

\paragraph{LVLM and Visual Grounding.} 

Building upon LLMs, LVLMs extend their capabilities to a multimodal context by incorporating visual perception into the generation process, with notable models such as GPT-4o \cite{openai2024gpt4ocard}, LLaVA \cite{liu2024visual}, Qwen2-VL \cite{wang2024qwen2vlenhancingvisionlanguagemodels}, and many others \cite{dai2023instructblip, zhu2024minigpt, chen2024far}, demonstrating advanced visual reasoning ability.
Additionally, some LVLMs employ grounding mechanisms to enhance multimodal interaction by allowing the model to reference specific regions of an image. 
This visual grounding has been achieved though the prediction of bounding boxes coordinates, as seen in models such as Kosmos-2 \cite{peng2023kosmos}, Shikra \cite{chen2023shikra}, BuboGPT \cite{zhao2023bubogpt}, Ferret \cite{you2024ferret}, Qwen2-VL \cite{wang2024qwen2vlenhancingvisionlanguagemodels}, and Groma \cite{ma2024groma}. 
To obtain a fine-grained localization of objects and semantic concepts pixel-level grounding has subsequently proposed in models such as Llava-Grounding \cite{zhang2024llava}, GLaMM \cite{rasheed2024glamm}, and GROUNDHOG \cite{Zhang_2024_CVPR}.
Unlike other grounding methods which learn to treat bounding box coordinates as part of the “language” of the model like Kosmos-2 and Shikra, or GROUNDHOG that was trained to output mask along with generated text, FiVL’s training process explicitly ties image tokens to their vocabulary text representation, creating fine-grained alignment that enhances visual grounding beyond simple region prediction, while preserving a text-based interface. The resulting model not only improves accuracy but can also be utilized to produce segmentation maps.

\paragraph{Visually Grounded Datasets.}
Training LVLMs require large-scale visual instruction-following data \cite{liu2024visual}. 
However, these datasets focus on the task of visual and language reasoning and generally do not have fine-grained image segmentation annotations. 
Prior work have mainly constructed custom datasets to train their respective grounded LVLM models.
In \cite{ma2024groma}, a custom dataset, Groma Instruct, was constructed by prompting GPT-4V to generate grounded conversations based on 30K samples with region annotations from COCO \cite{mscoco} and VG \cite{krishna2017visual}.
Llava-Grounding \cite{zhang2024llava} curated the Grounded Visual Chat (GVC) dataset by matching class labels of ground truth bounding boxes from COCO to noun phrases in conversations from LLaVA-Instruct-150K \cite{liu2024visual} using GPT-4.
The Grounding-anything Dataset (GranD) was specifically constructed to train GLaMM \cite{rasheed2024glamm} and utilized an object detection model to obtain visual entities that were then used to generate grounded dense captions through an LLM.
A grounded visual instruction tuning dataset, M3G2, was proposed to train the GROUNDHOG model \cite{Zhang_2024_CVPR}. There, the authors curated a dataset consisting of 2.5M text-image pairs for visually grounded instruction tuning derived and augmented from 27 existing datasets. Unlike previous datasets that depend on bounding boxes or align only noun phrases (e.g. entity object), our framework allows the alignment of various crucial word types, including adjectives and verbs 
(Table~\ref{table:words_attributes}).

\paragraph{Evaluating Visual Grounding.} 
A variety of benchmarks have been developed to ensure that VLMs rely on visual content rather than textual biases. Traditional methods such as VQA-CP \cite{vqa_cp} and GQA \cite{gqa} modify data splits or balance question-answer distributions to penalize overreliance on language priors, while synthetic sets like CLEVR \cite{clevr} remove commonsense priors to force explicit visual reasoning. Recent approaches (POPE \cite{pope}, NaturalBench \cite {naturalbench}) introduce adversarial or carefully constructed examples that only a visually grounded model can solve.
Others evaluate model robustness using image or question perturbations, such as CSS \cite{css} that generates counterfactual samples by removing relevant nouns in the image or question and assigning new ground-truth answers, and CARETS \cite{caret} which blurs or masks irrelevant background regions to evaluate model consistency.
In contrast to these perturbation-based methods, which often rely on annotated datasets and complex object selection, FiVL is applicable to any dataset and complements related benchmarks such as FiVL-POPE. FiVL explicitly identifies key visual expressions in a question-answer pair, applies vision masks, to compute a Visual Reliance Score. This metric assesses both the model reliance on the image as well as how well a benchmark necessitates visual context for accurate question answering.
 

\section{FiVL Framework}
\label{sec:dataset}

\begin{figure*}[ht]
    \centering
    \includegraphics[width=\textwidth]{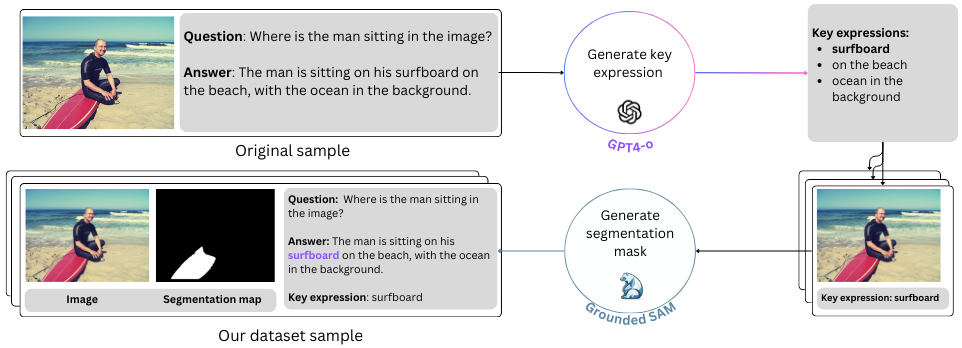}
    \caption{Dataset Collection Overview. First, GPT4-o processes the question and answer to produce "key expressions", which are then passed to GroundedSAM along with the image to produce segmentation maps. }
    \label{fig:datacollection}
\end{figure*}
In this section, we introduce our proposed framework, which offers two advantages over existing grounding datasets. Specifically, 1) our framework can augment any image-text dataset without relying on bounding box annotations, as these are generated on the fly, and 2) it enables fine-grained alignment with diverse types of textual content, extending beyond object entities as in prior work. We will then describe how our framework is utilized to generate both training and evaluation datasets.

\subsection{Data Collection Pipeline}
\label{subsec:dataset_pipeline}
We built grounded datasets for training and evaluation, by enhancing vision-question-answer and instruction datasets. Figure \ref{fig:datacollection} shows an overview of the pipeline. Each sample in the original datasets was augmented with key expressions, along with their corresponding bounding box indices and segmentation masks within the images as follows:
\paragraph{Key Expression Identification.}
The initial stage of data collection focused on identifying key expressions within each question-answer pair, using GPT-4o.
We refer to key expressions as specific words or phrases, like object names, attributes, or spatial relations, that rely on the visual context provided by the image.
We prompted GPT-4o with only the text of the question-answer pairs, omitting the images and asked it to detect essential expressions. The prompt is shared in Appendix \ref{prompt:instruct-system-prompt}. 
Using only questions and answers without visual cues allows GPT-4o to rely solely on linguistic context to determine whether certain words could be evoked based on text alone. This approach can help filter language-based answers from those needing visual context, while being computationally efficient. This process yielded a robust set of expressions, capturing the elements in each conversation that are closely tied to the visual information. 
\paragraph{Bounding Box and Segmentation Masks.}
To accurately associate key expressions with specific regions in each image, we used the GroundedSAM pipeline \cite{ren2024groundedsamassemblingopenworld}, which employs the GroundingDINO-tiny model \cite{liu2024groundingdinomarryingdino} for initial expressions localization generating bounding box indices, followed by the Segment Anything vit-huge model \cite{kirillov2023segany} for precise segmentation mask creation. Each key expressions was mapped to its relevant visual region, creating high-quality segmentation maps. If multiple segments corresponded to a single phrase, they were consolidated into a unified mask assigned to each token within the phrase, to maintain consistency across annotations. We removed segmentation mask of the same sample that overlapped by more than 95\%, ensuring that each segmentation map uniquely represents essential visual regions, avoiding redundancy and improving annotation clarity.


\subsection{Training Dataset}
\label{subsec:data_train}
Our training dataset, FiVL-Instruct, is built upon the LLaVA-1.5-mix-665K instruction tuning dataset \cite{liu2024improvedbaselinesvisualinstruction}, a vision-language instruction dataset containing 665K structured conversations between users and GPT. Most interactions begin with a user-provided image, followed by related questions, and GPT responses, each question-answer pair is referred as a turn.

We augmented the original LLaVA-1.5-mix-665K dataset by integrating the key expressions and their segmentation masks according to the pipeline outlined in Section \ref{subsec:dataset_pipeline}. 
Not every FiVL-Instruct sample includes a key expression. For such cases, we retained the original data point unchanged to maintain the dataset size for training. In our dataset, each conversation consists of multiple turns with an average of ten turns. 
Across the dataset, we collected 1.5 million unique segmentation masks for 2.3 million key expressions, averaging 2.3 masks and 3.5 key expressions per conversation. On average, a key expression consists of 2.4 words and the segmentation covers 28\% of the image. 
We also analyzed the types of key expressions in the resulting dataset. As shown in Table~\ref{table:words_attributes}, our expressions exhibit diverse types, making them distinct from those in prior grounding datasets.

\begin{table}[h!]
\centering
\resizebox{\columnwidth}{!}{%
\begin{tabular}{|c|c|c|c|c|c|}
\hline
Nouns & Adjectives & Proper Nouns & Adpositions & Verbs &  Others \\ \hline
42\% & 14\% & 10\% & 9\%  & 8\% & 17\%\\ 
\hline
\end{tabular}%
}
\caption{Statistics FiVL-Instruct dataset, showing key expressions words types.}
\label{table:words_attributes}
\end{table}
\subsection{Evaluation Datasets}
To assess the visual reliance of various LVLMs, we created three benchmark datasets derived from the following benchmarks: POPE \cite{pope}, VQAv2 \cite{vqav2}, and GQA \cite{gqa}. 

We selected these benchmarks,  because they each requires different levels of image reliance. POPE assesses sensitivity to visual perturbations, GQA evaluates understanding of detailed scene relationships, and VQAv2 tests visual grounding for diverse question types. Together, they offer a well-rounded assessment of how much models depend on visual information to answer accurately.
We followed the procedure outlined in Section \ref{subsec:dataset_pipeline} and produced FiVL-POPE, FiVL-VQAv2, and FiVL-GQA datasets.  To suit the nature of the evaluation datasets, we adapted the prompts for the key expression extraction (See Appendix \ref{appendix:system_prompts_keyexpress}, Figure \ref{prompt:eval_datasets}).
Unlike FiVL-Instruct, we filtered out samples without key expressions or segmentation maps resulting in a reduction in dataset sizes. As a result, FiVL-POPE covers 65\% of POPE, FiVL-VQA-v2 retains 40\% of VQA-v2 and FiVL-GQA accounts for 95\% of GQA size (refer Table \ref{table:evaldataset_size} in Appendix for the actual size of the filtered dataset numbers). 
This indicates that the original GQA relies more on visual context than POPE and VQAv2. Our evaluation sets: FiVL-POPE, FiVL-VQAv2, and FiVL-GQA, select subsets from the original datasets that require visual context and are better suited for visual-alignment testing.
Table~\ref{table:statistics_evaldataset} from Appendix presents additional statistics for these datasets. 



\section{Method Evaluation}
To ensure the quality of our framework, we conducted a multi-step evaluation process on the training dataset described in Section \ref{subsec:data_train}. This included both human-based evaluations and automated assessments, allowing us to validate the relevance and accuracy of the key tokens and their alignment with visual content. Below, we outline the key components of our evaluation strategy.
\subsection{Human Evaluation}
We conducted a manual evaluation in order to validate the coherency of the key expressions as well as the relevancy of the segmentation maps with respect to the formers. For each sample, we presented to the annotators one random key expression with its associated segmentation map. Annotators were asked three questions: \textit{whether the key expression aligns with the definition provided in Section \ref{subsec:dataset_pipeline}}, \textit{if the segmentation map is relevant to the key expression}, and \textit{whether the sample is of good quality} (does the text makes sense, is the answer related to the question).  In total, 557 unique samples were annotated by 12 different annotators. A screenshot of the API is shown in Appendix \ref{appendix:api}. 
Results show that 77\% of the annotators labeled the samples as overall good data points.
In the key expression evaluation, 75\% of key expressions were deemed pertinent. In the segmentation map evaluation, 58\% of segmentation map were annotated as relevant to the key expression. The last result can be explained by the fact that some key expressions might inherently be more abstract or complex or by the performance of the GroundedSAM pipeline. 
Finally, if we compute the key expressions and segmentations score only for the samples annotated as "good data point overall", 85\% of the data are with valid key expressions and 69 \% are with relevant segmentation masks.
Additionnaly, we find that the quality of the segmentation is related to its size. Figure \ref{fig:segmentation_mask_impact} from Appendix indicates that when the segmented mask occupies less than 20\% of the image, annotators were more likely to consider the segmentation relevant. To train our model (see Section \ref{subsec:results_train}), we selected the segmentation masks based on their size. This metric can be used as a threshold-based filtering method for future applications of the dataset.
\subsection{Automatic Evaluation}
 Inspired by recent applications using GPT4-o as-a-judge ~\cite{zheng2023judging}, we designed two prompting techniques to automatically assess the quality of extracted keywords and segmentation masks based on a given keyword. Both evaluations were conducted on a randomly sampled set of 1,957 keywords and their corresponding segmentations from FiVL-Instruct.

\subsubsection{Keyword Evaluation}
We prompt GPT4-o, prompts are presented in Appendix \ref{appendix:system_prompt_eval}, to evaluate the correctness of the key expressions and report the following metrics:
\textit{Importance Ratio = 76\%} representing the percentage of extracted expressions classified as key expressions. This result is close to human evaluation, which is 75\%. \textit{Overall Importance Degree = 6.8}, which indicates the average importance score across all keywords, regardless of GPT-4o classification. And, \textit{Importance Degree of Important Keywords = 9.0}, which calculates the average importance ratio of keywords identified as important by GPT-4o. 
These metrics indicate the high quality of our keywords.


\subsubsection{Segmentation Evaluation}
Given a keyword, we aim to evaluate whether our segmentation for this keyword is accurate. We designed two prompts to assess the quality of the segmentation: first, we check if the segmentation content adequately covers the keyword (Seg1); second, we verify that the inverse of the segmentation does not contain any content related to the keyword (Seg2). 
Both prompts are given in Appendix \ref{appendix:system_prompt_eval}. 
Results show that only for \(Seg1=46\%\) of the cases GPT-4o capture the keywords in the segmentation.  
On one hand, this result aligns with the manual annotations and can be addressed in the same manner. 
On the other hand, we found that segmentations classified as good often involve specific objects (e.g., tennis players, bears). In contrast, segmentations classified as bad are often abstract concepts (e.g., water pressure, mental game, splashing), descriptive words (e.g., unique, uneven ground), or complex actions (e.g., walking over logs). These types of words are difficult to link to a specific part of an image when the full image context is not provided. This also highlights the limitations of the first type of evaluation prompt.
In \(Seg2=72\%\) of cases, the model determines that the inverse of the segmentation is irrelevant to the keywords, accurately recognizing that without the segmented mask, the key expressions are not present in the image. This measures if we do not miss key objects in our segmentation maps. If 2 objects appear in the image not at the same positions, we make sure that our maps contain both of them.

\section{Applications of FiVL Datasets}
In this section, we describe three approaches to utilize our datasets. 
Section \ref{subsec:results_train} describes how FiVL can be used as a training dataset and the resulting models not only achieve better performance but also has one more capability than the baseline model: generate segmentation maps. Section \ref{subsec:results_eval} introduces FiVL as a tool for evaluating the visual reliance of LVLMs. Section \ref{sec:explainability}
shows that FiVL can assist the interpretability of models.

\subsection{Training}
\label{subsec:results_train}
We introduce here a training task referred to as Vision Modeling. To assess the effectiveness of this task, we fine-tuned an LVLM, specifically, LLaVA-1.5-7b \cite{liu2023visualinstructiontuning}, referred as to the baseline, on FiVL-Instruct.
For training our model, we used only key expressions that appeared verbatim in the answers for each turn, focusing exclusively on noun-based key expressions.

\paragraph{Method.}
In the original LLaVA training,  it has two stages: the first pretraining stage trains a projector which aims to align visual and textual representations, while the second finetuning stage performs only language modeling on the textual outputs of the LM head. In this work, we propose to guide the visual outputs of the last linear layer during the finetuning stage, in addition to performing language modeling on its textual outputs. 
Our approach augments the Language modeling cross-entropy loss with a Vision modeling (VM) cross-entropy loss where each patch that belongs to a segmentation map is trained to predict the related keyword from the vocabulary.

We denote by \( x \) the input and \( y \) the logits with respect to each token. The logits are the outputs of the last linear layer that projects the last hidden states to the vocabulary space:

\begin{align*}
    x &= (x_{i_0}, x_{i_1}, \dots, x_{i_{Ni}}, x_{t_0}, x_{t_1}, \dots, x_{t_N}), \\
    y &= (y_{i_0}, y_{i_1}, \dots, y_{i_{Ni}}, y_{t_0}, y_{t_1}, \dots, y_{t_N}),
\end{align*}

where \(N_{i}\) is the number of image tokens, \(N_{t}\) the number of text tokens, \(N\) the total lenght. \( x_i \) are the inputs embedding that relate to the image tokens and \( x_t \) to the text tokens;  \( y_i \in \mathbb{R}^{N_{i} \times \text{vocabulary\_size}} \) represents visual logits, while \( y_t \in \mathbb{R}^{N_{t} \times \text{vocabulary\_size}} \) represents textual logits.

\begin{figure}
\centering     
{\includegraphics[width=\columnwidth]{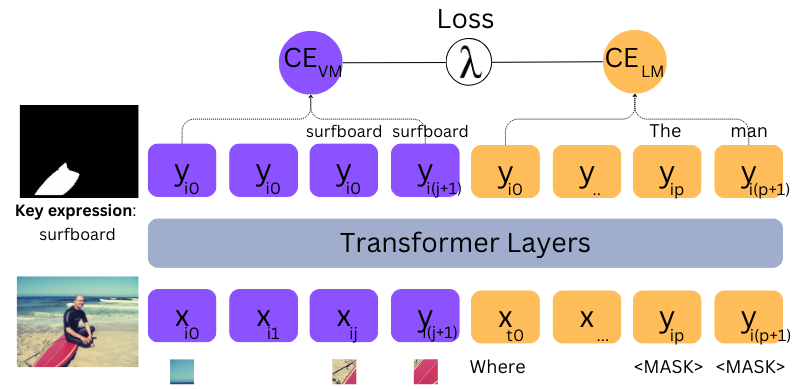}}
\caption{Overview of Vision Modeling pretraining task.}
\label{fig:visalog_overview}
\end{figure}

In Language Modeling (LM), only \( y_t \) related to the answer are trained. We propose to also train \(y_i\) related to the segmented piece. Figure \ref{fig:visalog_overview} shows an example where given a picture, a question,  the LM loss would only guide the relevant tokens \( y_t \) to be the expected answer \textit{The man is sitting on his surfboard <...>}. In our method, we also do vision modeling by training each visual logit corresponding to the segmented mask to refer to the noun from the key expression: \textit{surfboard} from the text vocabulary. 
In order to create the vision labels we proceed like such: for each sample, each image token will be assigned to exactly one token in the text vocabulary. The selection is based on the size of the mask (we take the smallest) and the type of the keyword (we filter only nouns). 
That way, for each image patch, there is maximum one key token that describes the patch. Image patches that do not have a related keytoken are ignored in the loss, similar to LM. We then compute a weighted sum from the cross-entropy, \( CE_{VM} \) between the created vision labels and the visual logits and the cross-entropy related to language modeling,  \( CE_{LM} \).
The resulting loss is computed as such:
\begin{equation}
\label{eq:loss}
    L = \lambda * CE_{VM} +  (1-\lambda) * CE_{LM}, \lambda \in [0,1]
\end{equation}

\begin{figure}[tbh]
\centering     
{\includegraphics[width=\linewidth]{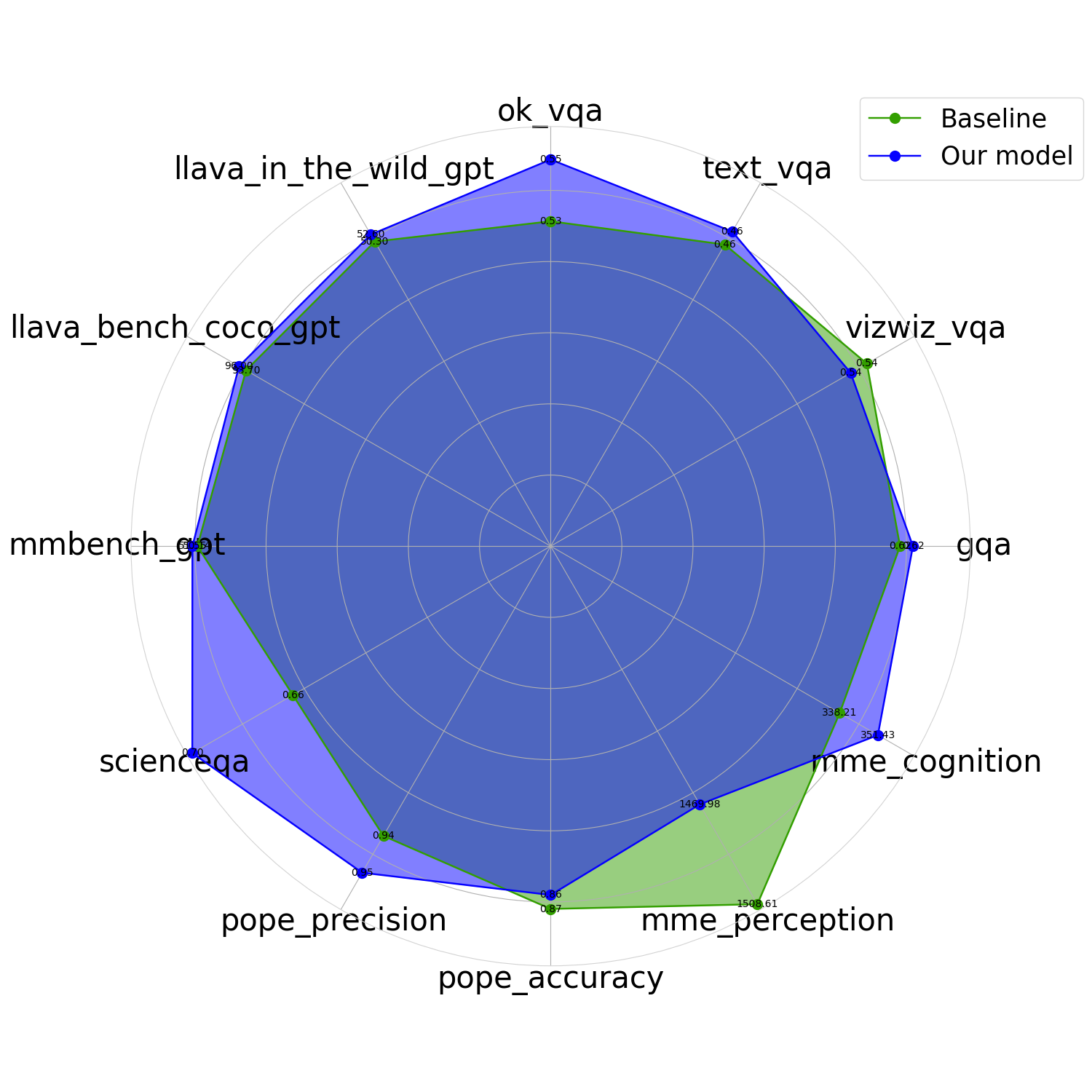}}
\vspace{-2em}
\caption{Our model trained on FiVL-Instruct evaluated on various benchmarks compared to the baseline.}
\label{fig:visavm_results}
\end{figure}

\paragraph{Improve Benchmark Results.}

We conducted multiple experiments to determine the optimal hyperparameters. We finetuned LLaVa-v1.5-7b from scratch using our augmented dataset. We used the trained multimodal projector and started from Vicuna-v1.5-7b \cite{zheng2023judgingllmasajudgemtbenchchatbot} weights. We maintained the original training setup (batch size, number of epochs, etc.) and primarily focused on experimenting with different learning rates and \( \lambda \). The best results were achieved with a learning rate of 2-e5, the same as in the original setup, and  \( \lambda \) set to 0.1. Training details are shared in Appendix \ref{appendix:training_details} and ablations are reported in Appendix \ref{appendix:ablations}. Figure \ref{fig:visavm_results} shows how we outperformed the baseline in different benchmarks: OK-VQA \cite{marino2019okvqavisualquestionanswering}, MME \cite{fu2024mmecomprehensiveevaluationbenchmark}, POPE \cite{pope}, ScienceQA \cite{lu2022learnexplainmultimodalreasoning}, MMBench \cite{liu2024mmbenchmultimodalmodelallaround}, LLaVA-Bench-COCO \cite{liu2023visualinstructiontuning}, LLaVA-in-the-wild \cite{liu2023visualinstructiontuning}, Text-VQA \cite{singh2019vqamodelsread}, VizWiz-VQA \cite{gurari2018vizwizgrandchallengeanswering}, GQA \cite{gqa}.

\begin{figure}[t]
    \centering
    

    \begin{minipage}{0.20\textwidth}
        \centering
        \includegraphics[width=\linewidth]{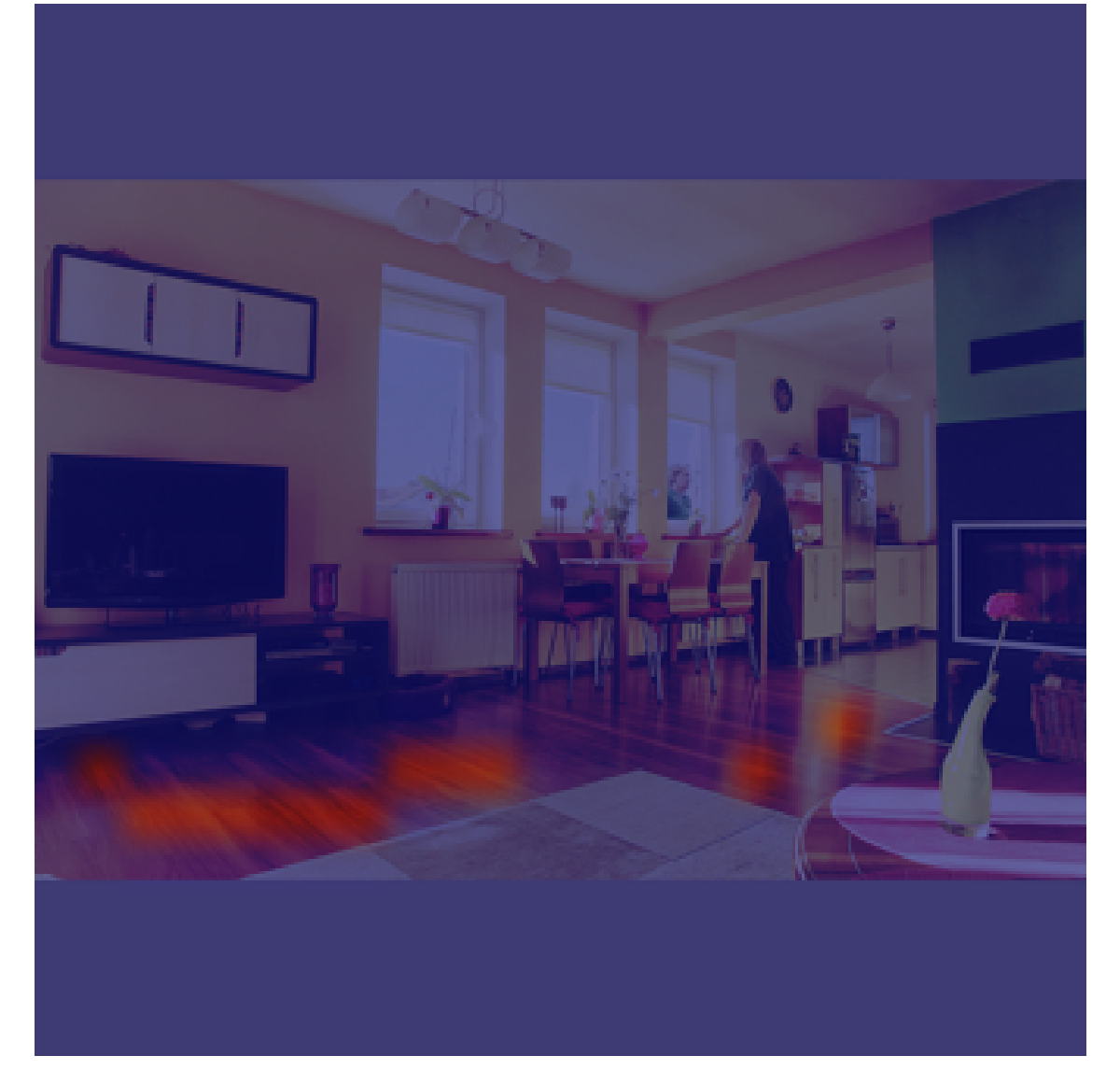 }
        \subcaption{Baseline}
    \end{minipage}
    \hfill
    \begin{minipage}{0.20\textwidth}
        \centering
        \includegraphics[width=\linewidth]{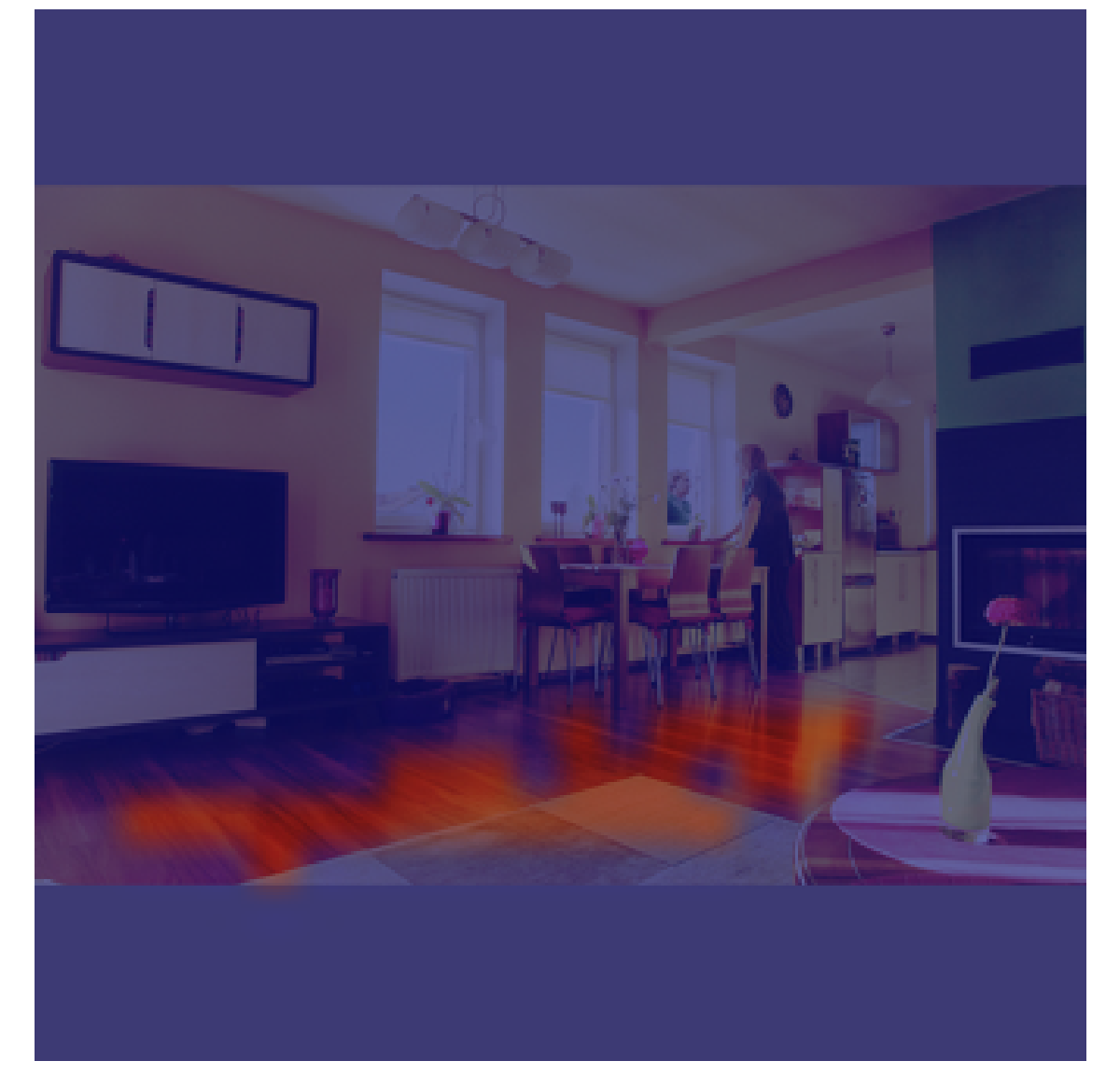}
        \subcaption{Our model}
    \end{minipage} 
    \caption{Predicted token from vision logits ( "Flo", for "floor") and its corresponding regions in the image.}
\label{fig:segmentation_vm}
\vspace{1em}
\end{figure}

\paragraph{Better Grounding Outcome.}
Figure \ref{fig:segmentation_vm} compares the baseline model and FiVL, illustrating the correspondance of each image patch with its related most probable token from the vocabulary. The argmax of the vision logits is identified, mapped back to the text vocabulary. Then, for each token, the relevant image patches are highlighted, indicating which parts of the image align with that token. Although the baseline can capture some relevant text tokens for the image patches and tends to scatter semantically similar image patches across different tokens from the vocabulary. Some of these tokens may be relevant, but others are not, indicating a lack of consistent grounding. On the other hand, our model shows more relevant images patches related to the word. 

\paragraph {Vision Logits as Approximate Segmentation Maps.} 
Another interesting finding is that we can obtain a weak ``segmentation maps" by predicting the most probable text tokens from the vision logits. As a simple observation, averaging over 100 examples, the baseline predicts 74 different tokens overall (with lots of unrelated tokens such as "a", "*", "is" etc.), while our model only encompasses 9 tokens. 
This demonstrates potential in leveraging visual logits for segmentation.
as shown in Figure \ref{fig:segmentation_vm}. 
More examples are presented in Figure \ref{fig:segmentation_images} in Appendix \ref{appendix:segmentation_analysis}.
We further conducted evaluations to assess the performance of the segmentation capability of our model. Results, reported in Appendix \ref{appendix:segmentation_analysis}, shows FiVL's enhanced ability to produce precise and coherent segmentation masks. 

\subsection{Visual Reliance Evaluation} 
\label{subsec:results_eval}
FiVL datasets also allow us to measure \textit{Visual Reliance} by performing perturbation based evaluation: first assessing model accuracy on the original images, then on the masked images. We introduce a \textit{Visual Reliance Score} in Eq.\ref{eq:visual_reliance}, which measures the percentage of drop in accuracy from the original to the masked image version. A higher score indicates stronger model dependency on visual input. 

{\small 
\begin{equation}
\text{Visual Reliance Score} = 
\frac{\text{accuracy}_{\text{original}} - \text{accuracy}_{\text{perturb}}}
{\text{accuracy}_{\text{original}}}
\label{eq:visual_reliance}
\end{equation}
}
Indeed, the perturbation based on the masked image is not perfect, it still provides a measurement of visual reliance. To confirm that FiVL is suitable for evaluating visual reliance, we created a control dataset with random masking. In this control set, each image contains a bounding box mask of the same size as the key expression mask, but placed at a random location within the image. This approach provides a comparison to determine whether performance declines specifically due to masking critical visual areas or simply from general occlusion.

We compared the performance of two models, LLaVA-v1.5-13b and Qwen2-VL-7B-Instruct \cite{wang2024qwen2vlenhancingvisionlanguagemodels}, across the three evaluation datasets we created.

\begin{table}[b]
\centering
\resizebox{\columnwidth}{!}{%
\begin{tabular}{|c|c|c|c|c|c|c|}
\hline
 & \multicolumn{2}{c|}{VQA-v2} & \multicolumn{2}{c|}{GQA} & \multicolumn{2}{c|}{POPE} \\
\hline
 & FiVL & Random & FiVL & Random & FiVL & Random \\
\hline
LLaVA-13B & 0.72 & -0.05 & 0.33 & 0.03 & 0.49 & 0.02 \\
\hline
Qwen2-VL-7B & 0.64 & 0.07 & 0.38 & 0.03 & 0.47 & 0.02 \\
\hline
\end{tabular}%
}
\caption{Comparison of Visual Reliance Score between FiVL bounding boxes and random perturbations across benchmarks and models.}
\label{table:drops}
\end{table}

\paragraph{Compare Perturbation Methods.} Table~\ref{table:drops} compares FiVL and Random Perturbation. It shows that across all benchmarks and models, the perturbation based on FiVL masks causes a significantly larger performance drop compared to random perturbation. This indicates that our bounding boxes capture meaningful visual content relevant to the questions and FiVL represent good testbeds for visual reliance. 


\paragraph{Compare Models and Benchmarks.} 
To gain a broader understanding of model/benchmark performance, we evaluated five models on FiVL-VQAv2, FiVL-POPE, and FiVL-GQA. This helps to assess the generalizability of our approach across more models. Table \ref{table:drops_appendix} shows our results for Qwen2-VL-7B, LLaVA-v1.5-7b\cite{liu2023visualinstructiontuning}, LLaVA-13B, GPT4o \cite{openai2024gpt4ocard}, BLIP-2 \cite{li2023blip2bootstrappinglanguageimagepretraining}, Pixtral-12B \cite{agrawal2024pixtral12b} and Phi3-Vision\cite{abdin2024phi3technicalreporthighly}, which are state-of-the arts mutlimodal models. In bold, are the highest visual reliance scores per model and across all benchmarks.
The results unanimously indicate that, among all models, FiVL-VQAv2 requires models to rely on the image the most compared to other datasets. 
Underlined are the highest visual reliance scores across models, given a benchmark. 
Looking at the average performance per model across benchmarks (last column), we observe that GPT4-o relies most heavily on the image as a reference for answering, followed by Pixtral-12B. 
Lastly, we observe a correlation between overall model performance and the Visual Reliance score. 
According to the available VLM Leaderboard \cite{open_vlm_leaderboard}, that measures the performance of the models on a broad range of benchmarks, and Table \ref{table:drops_appendix}, we see that GPT4o (ranked 20 on the leaderboard) has a higher overall Visual Reliance Score compared to Pixtral-12B (ranked 54). Within a similar Visual Reliance Score range, follow LLaVA-13B (118), Qwen2-VL-7b (136) and Phi3-V (77). Lastly, LLaVA-7b (127) appears at a lower rank. This suggests that this average of Visual Reliance Scores captures the overall performance of the model and is not overly sensitive to the specific benchmarks used. All together, these results indicate that effective image utilization is a key factor in achieving higher performance.

\begin{table}[h]
\centering
\resizebox{\columnwidth}{!}{%
\small
\begin{tabular}{|c|c|c|c|c|}
\hline
  & VQA-v2 & GQA & POPE & Avg VRS \\ 
  \hline
Qwen2-VL-7B & \textbf{0.64} & 0.38 & 0.47  & 0.50 \\ \hline
LLaVA-13B & \textbf{0.72}& 0.33 & 0.49  & 0.51 \\ \hline
LLaVA-7B & \textbf{0.56} & 0.31 & 0.47 & 0.45 \\ \hline
GPT4o & \textbf{0.74} & \underline{0.63} & 0.49 &  \underline{0.62} \\ \hline
BLIP-2 & \textbf{0.52} & 0.23 & 0.03 & 0.26 \\ \hline
Pixtral-12B & \underline{\textbf{0.75}} & 0.58 & 0.42 & 0.58 \\ \hline
Phi3-V & \textbf{0.60} & 0.33 & \underline{0.54} & 0.49 \\ \hline
Avg & \textbf{0.65} & 0.40 & 0.42 & - \\ \hline
\end{tabular}%
}
\caption{Visual reliance scores (VRS): \% of drop in performance using FiVL bounding boxes for perturbations. In \textbf{bold}: highest scores across benchmarks. \underline{Underlined}: highest scores across models. Avg stands for Average. }
\label{table:drops_appendix}
\end{table}

\subsection{Explainability}
\label{sec:explainability}
\begin{figure}[t!]
  \centering
  \begin{subfigure}[t]{0.22\textwidth}
    \centering
    \includegraphics[width=\textwidth]{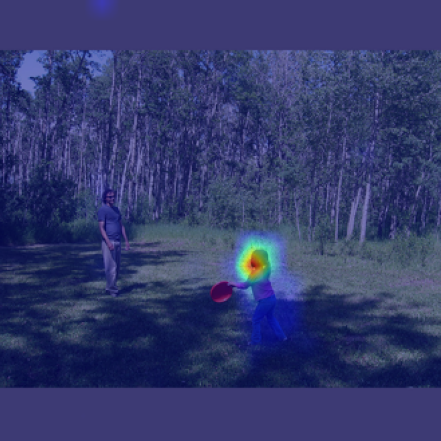}
    \caption{Attention Head (10,6)}
    \label{fig:head_10_6}
  \end{subfigure}
  \quad
  \begin{subfigure}[t]{0.22\textwidth}
    \centering
    \includegraphics[width=\textwidth]{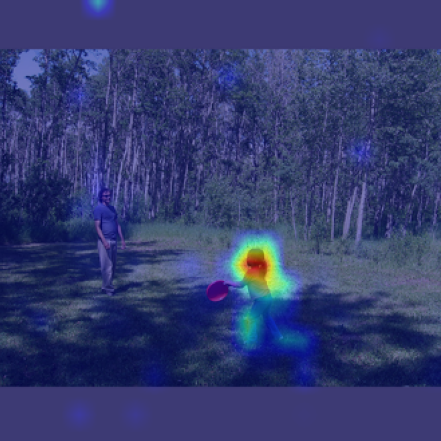}
    \caption{Attention Head (14,11) }
    \label{fig:head_14_11}
  \end{subfigure}

  \vspace{0.5em} 

  \caption{
  Attention heatmaps overlaid on the original images for attention heads (10,6) and (14,11) of the token "girl" for the answer: \textit{The two people  [...] are a man and a little \textbf{girl}}.}
  \label{fig:xai}
\end{figure}

We show that FiVL can assist the interpretability of LVLMs by generating a summary plot showing a vision-alignment metric computed across all heads and layers, as introduced in \cite{aflalo2022vlinterpretinteractivevisualizationtool}.  Using Spearman correlation between the segmentation mask of FiVL-Instruct dataset and the attention to the corresponding key expression tokens in the Vision-to-Language attention component, we are able to retrieve the heads achieving the strongest VL alignment. The head summary (Appendix \ref{appendix:xai}, Figure \ref{fig:head_summary}) indicates that heads (10,6) and (14,11) are effective at aligning vision with language. For instance, Figures \ref{fig:head_10_6} and \ref{fig:head_14_11} show from which patches of the image the token \textit{girl} gets the most attention, clearly focusing on the girl.




\section{Conclusion}

In this paper, we introduced FiVL, a framework designed to enhance vision-language alignment in large vision-language models. We applied our approach across key stages of an LVLM training workflow: training, evaluation, and explainability. By training a LLaVA model using the FiVL dataset and our novel training task, we measured improvement in a majority of benchmarks and produced a built-in feature that segments the image. Our evaluation datasets measured model reliance on images for answering questions, offering insights into the level of image dependency required across benchmarks. The results indicate a correlation between this dependency and overall model performance.
Finally, our explainability application enables users to identify attention heads that excel in vision-language alignment, allowing for a deeper understanding of potential hallucinations. 

\section*{Limitation}


 In this work, we utilize the FiVL framework to augment LLaVA instruction fine-tuning data and train a new model to compare against the baseline, demonstrating the effectiveness of our proposed framework and training objectives. However, we have only investigated LLaVA model, because of the limited availability of open-source training datasets for other LVLMs and augmenting additional data incurs additional inference costs.
Lastly, we rely on an off-the-shelf segmentation model (GroundedSAM) that takes a simple text prompt and an image as input. In our context, this may lead to less accurate segmentation, as the full contextual understanding of keywords might be necessary. To mitigate this issue, we could apply a filtering technique to enhance the overall quality of the dataset.

\bibliography{custom}
\clearpage
\appendix

\section{Appendix}
\label{sec:appendix}
\appendix
\section*{Appendix}
\section{Evaluation dataset}
Table \ref{table:evaldataset_size} compares the size of our datasets with the original datasets and Table \ref{table:statistics_evaldataset} presents some statistics of FiVL-VQA-v2, FiVL-GQA and FiVL-POPE.

\begin{table}[h!]
\centering
\resizebox{0.8\columnwidth}{!}{%
\begin{tabular}{|c|c|c|c|}
\hline
 & VQA-v2 & GQA & POPE \\ \hline
Original & 9,999 & 12,280 & 9,000 \\ \hline
FiVL & 4,040 & 11,660 & 5,870 \\ 
\hline
\end{tabular}%
}
\caption{Evaluation dataset sizes after filtering out samples without key expressions or segmentation masks.}
\label{table:evaldataset_size}
\end{table}

\begin{table}[h!]
\centering
\resizebox{\columnwidth}{!}{%
\begin{tabular}{|c|c|c|c|}
\hline
 & FiVL-VQAv2 & FiVL-GQA & FiVL-POPE \\ \hline
Key expressions & 1.27 & 1.5 & 1\\ \hline
Segmentation masks & 3.79 & 4.71 & 3.48 \\ 
\hline
\% of masked pixels & 24\% & 21\% & 16\% \\ \hline
\end{tabular}%
}
\caption{Statistics per sample of our evaluation datasets. First row details the average number of key expressions, second row describes the average number of distinct segmentation masks and last row describes the average percentage of the pixels that were masked. }
\label{table:statistics_evaldataset}
\end{table}

\begin{figure}[h!]
    \centering
    \includegraphics[width=\columnwidth]{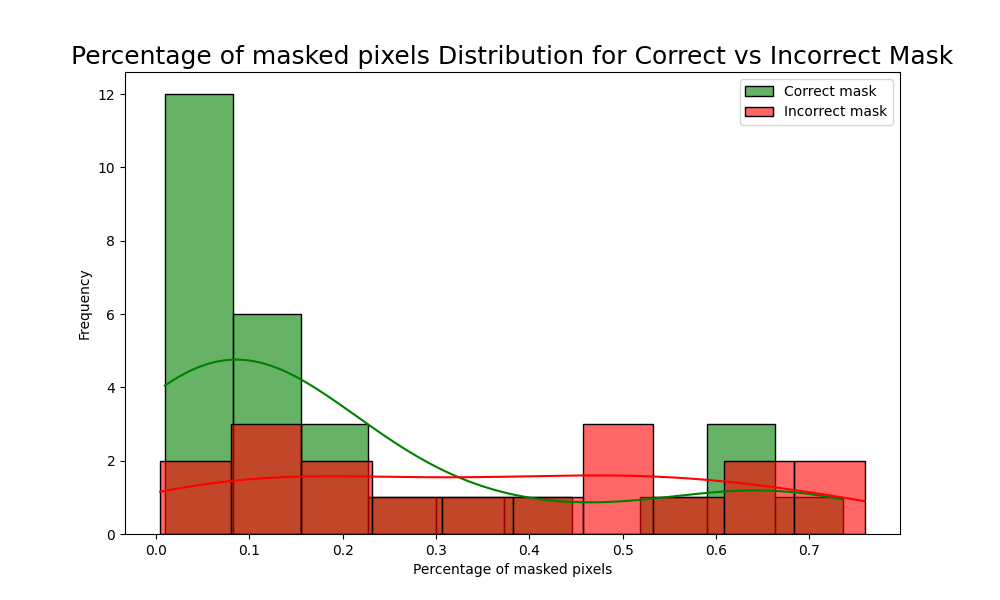}
    \caption{Impact of the size of the segmentation mask. Comparison of the Percentage of masked pixels Distributions for Correctly and Incorrectly annotated Masks}
    \label{fig:segmentation_mask_impact}
\end{figure}

\section{System prompts for key expressions retrieval}
\label{appendix:system_prompts_keyexpress}
We use GPT-4o via the Azure OpenAI API to extract the key expressions of the datasets we considered. In this section, we share the prompts used for this step of the data collection.
We had to use slightly different prompts for the training datasets compared to the evaluation datasets. In the training datasets, where instructions are open-ended question-answer pairs, the key expressions are often found in the answer. However, in the evaluation datasets, we encountered questions that required specific types of responses (yes/no questions, counting etc...). In these cases, the key expressions are typically found in the question instead. For references, we have provided the prompt used for training dataset in Figure~\ref{prompt:instruct-system-prompt} and prompts used for evaluation datasets VQA-V2, GQA, and POPE in Figure~\ref{prompt:eval_datasets}. For each benchmark we use different examples that suit the best to the types of questions. See Figure \ref{prompt:ex_vqa} for FiVL-VQAv2 and Figure \ref{prompt:ex_gqa_pope} for FiVL-GQA and FiVL-POPE.

\section{System prompts for evaluation}
\label{appendix:system_prompt_eval}
We used GPT-4o as LLM-as-a-judge in order to evaluation the correctness of the key expressions and the segmentation maps. The system prompt are shared in Figure \ref{fig:seg_prompt} and Figure ~\ref{fig:keyword_prompt}.

\begin{figure}[h]
\lstset{frameround=fttt}
\begin{lstlisting}[frame=trBL,linewidth=1.01\columnwidth,breaklines=true,breakautoindent=false,breakindent=0pt,numbers=none,basicstyle=\ttfamily\footnotesize]
[Seg1] You are given a part of the image and a word/phrase, do you think this is a good segmentation that the given part of the image covers this word/phrase?

Word/phrase: {word}

Answer only "yes" or "no".

[Seg2] You are given a part of the image and a word/phrase, do you see any part of the image that is related to the word? 

Word/phrase: {word}

Answer only "yes" or "no".

\end{lstlisting}
    \caption{Segmentation Verification Prompt for GPT-4o.}
    \label{fig:seg_prompt}
\end{figure}

\begin{figure}[H]
\lstset{frameround=fttt}
\begin{lstlisting}[frame=trBL,linewidth=1.01\columnwidth,breaklines=true,breakautoindent=false,breakindent=0pt,numbers=none,basicstyle=\ttfamily\footnotesize]
You are given a question, a word/phrase and an image. Please rate the importance degree from 0-10 scale ([OID]). 
Note that
 - 0 means not important at all and 10 means very important. 
 - Important word/phrase means that this word/phrase is closely related to the image and the question, and it could not be evoked without the use of the image (IR).
 - If the question does not related to the image, in other words, the answer does not depend on the image content, then any words are not important. 

Question: {question}

A word: {word} 

Only answer important or not important, and the importance degree from 0-10?

\end{lstlisting}
    \caption{Keyword Verification Prompt for GPT-4o.}
    \label{fig:keyword_prompt}
\end{figure}

\section{Explainability}
\label{appendix:xai}

\begin{figure}[t!]
    \centering
    \includegraphics[width=0.9\linewidth]{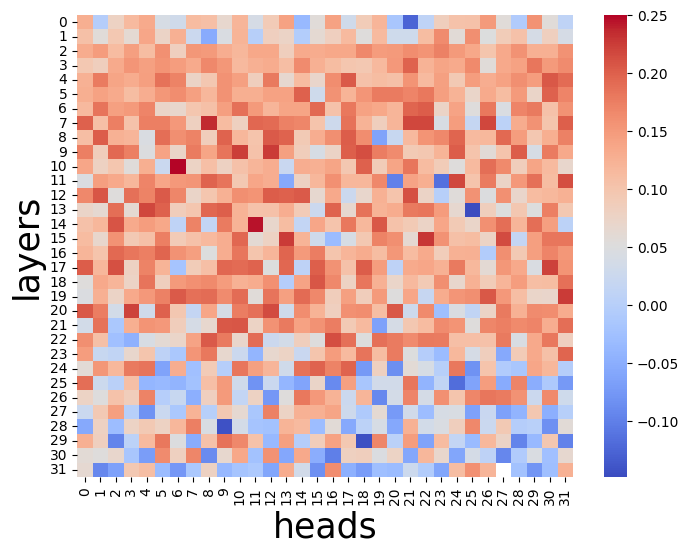}
            \caption{Head summary for VL alignment via Spearman correlation between token segmentation and vision attention}
        \label{fig:head_summary}
\end{figure}

For an attention matrix of size
\( (N_{layers}, N_{heads}, N_{i} + N_{t}
, N_{i} + N_{t}) \), The head summary calculates the statistical mean over the last two dimensions, producing a plot with dimensions of \( (N_{layers}, N_{heads}) \) averaged for 500 samples. For a given question, image, key expression and related segmentation mask from the FiVL-Instruct dataset, we generate the answer using LlaVA-v1.5-7b. We then identify if the key expression is in the answer or in the question. If so, we probe each head by computing the Spearman correlation between the segmentation mask \( ( \sqrt{N_{i}},\sqrt{N_{i}} ) \). and the attention to the corresponding key expression tokens in the Vision-to-Language attention component \( (1, 1, N_{i}
, 1 ) \) (first dimension selects the layer, second the head and the last dimension corresponds to the key token) for each head. This is performed on the language model component but not on the vision component of LLaVA.
In this way, we identify the attention heads that ground the most the two modalities by performing a function similar to object segmentation. Figure \ref{fig:head_summary} shows the head summary and the corresponding language-vision attention weights related to the key expression tokens displayed as a heatmap over the image. The head summary shows that the heads achieving the strongest vision-language alignment are in the early layers. This might be due to the fact that the input to this transformer is the output of multimodal projector of LLaVA, which is designed specifically to align these two modalities. The head summary indicates that heads $(10,6)$ and $(14,11)$ are effective at aligning vision with language. 
In this way, we identify the attention heads that ground the most the two modalities by performing a function similar to object segmentation. Figure \ref{fig:head_summary} shows the head summary and the corresponding language-vision attention weights related to the key expression tokens displayed as a heatmap over the image. The head summary shows that the heads achieving the strongest vision-language alignment are in the early layers. This might be due to the fact that the input to this transformer is the output of multimodal projector of LLaVA, which is designed specifically to align these two modalities. The head summary indicates that heads (10,6) and (14,11) are effective at aligning vision with language.

\begin{figure}[t!]
  \begin{subfigure}[t]{0.22\textwidth}
    \centering
    \includegraphics[width=\textwidth]{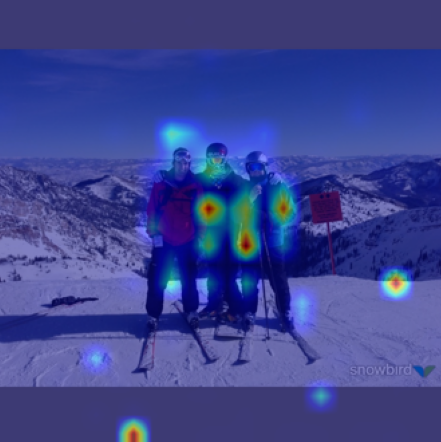}
    \caption{Attention Head (10,6) of the token \textit{three}. A - \textit{There are \textbf{three} people in the image}.}
  \end{subfigure}
  \quad
  \begin{subfigure}[t]{0.22\textwidth}
    \centering
    \includegraphics[width=\textwidth]{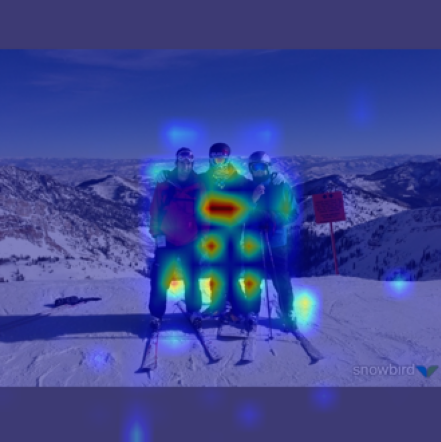}
    \caption{Attention Head (14,11) of the token \textit{three}. A - \textit{There are \textbf{three} people in the image}.}
  \end{subfigure}

  \caption{
  Attention heatmaps overlaid on the original images for attention heads (10,6) and (14,11), which have a high Spearman correlation, to probe vision-language alignment. }
  \label{fig:xai}
\end{figure}

\section{Training details}
\label{appendix:training_details}
To train our model, we used 8 Nvidia RTX A6000 GPUs using the hyperparameters from Table \ref{table:hyperp}

\begin{table}[ht]
\centering
\begin{tabular}{rrl}
    \toprule
    & Batch Size & $4$ \\
    & Number of GPUs & $8$ \\
    & Gradient Accumulation & $4$ \\
    & Number of epochs & $1$ \\
    & LLaVA Image Size & $576$ \\
    & Optimizer & SGD \\
    & Learning Rate & $2e-5$ \\
    & $\lambda_{VM}$ & $0.1$ \\
    & BF16 & $True$ \\
    & LR scheduler & $cosine$ \\
    & Vision Tower & openai/clip-vit-large-patch14-336 \\
    & Language Model & lmsys/vicuna-7b-v1.5 \\
    \bottomrule
\end{tabular}
\caption{Hyperparameters to train our model.}
\label{table:hyperp}
\end{table}

\section{Ablations}
\label{appendix:ablations}
\begin{figure}[htbp]
    \centering
    \begin{subfigure}[b]{0.5\textwidth}
        \centering
        \includegraphics[width=\textwidth]{images/ablations_fivl_Merged_dataset.png}
        \caption{Ablations on the dataset size}
        \label{fig:ablation_datasetsize}
    \end{subfigure}
    \begin{subfigure}[b]{0.5\textwidth}
        \centering
        \includegraphics[width=\textwidth]{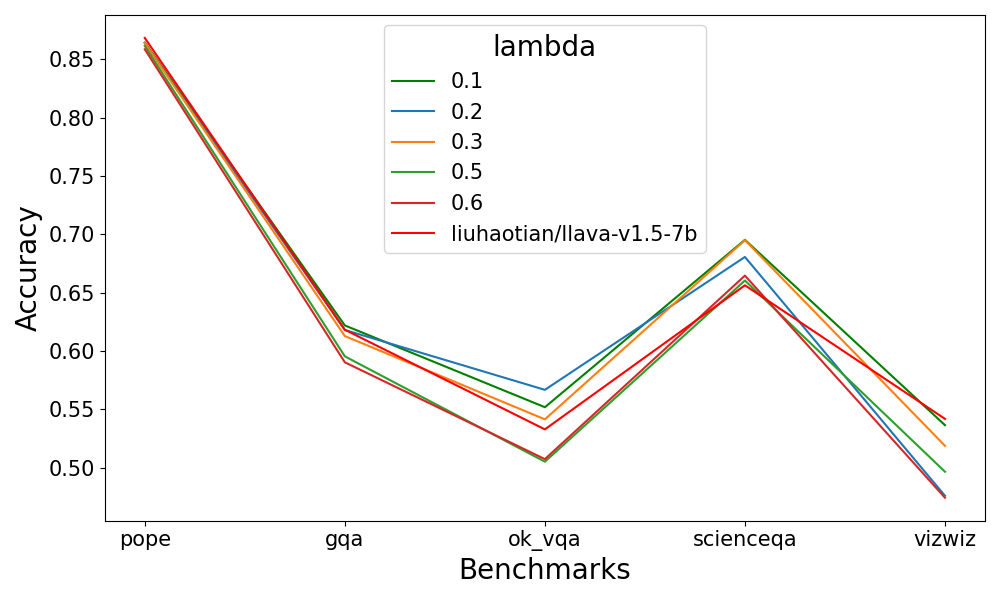}
        \caption{Ablations on the $\lambda$ parameter}
        \label{fig:ablation_lambda}
    \end{subfigure}
        \begin{subfigure}[b]{0.5\textwidth}
        \centering
        \includegraphics[width=\textwidth]{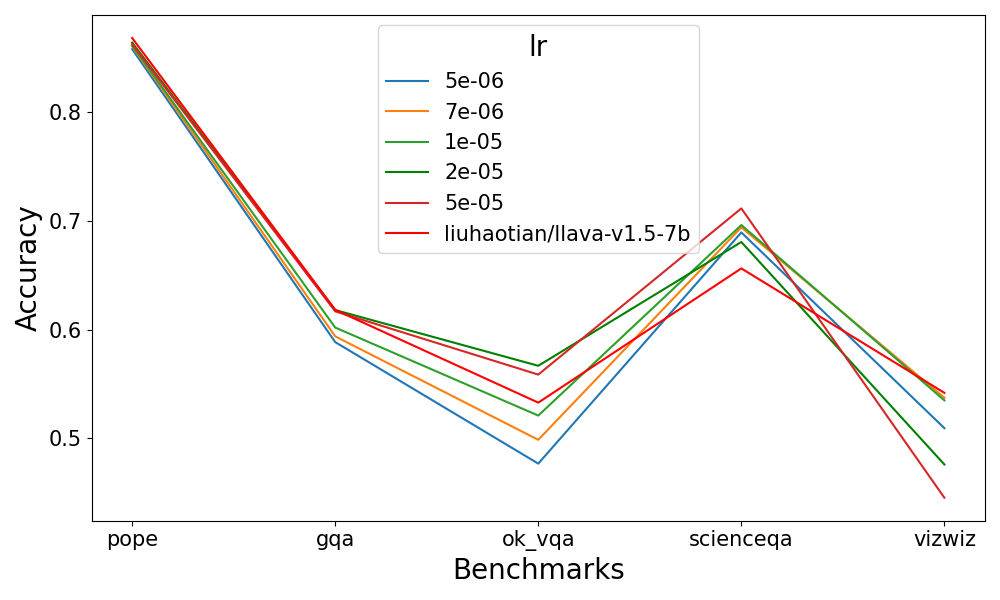}
        \caption{Ablations on the learning rate}
        \label{fig:ablation_lr}
    \end{subfigure}
    \caption{Ablations for the training method}
    \label{fig:main_figure}
\end{figure}

We conducted ablations studies on different parameters of the model. 
In this section, we will limit the experiments on a subset of the benchmarks. 
As mentioned in Section \ref{sec:dataset}, the FiVL-Instruct dataset includes some samples without key expressions. We first trained our model using only on the samples that had at least one associated key expression and segmentation map. We conducted another experiment by merging the remaining samples with these. Figure \ref{fig:ablation_datasetsize} presents the results of this ablation, showing that merging the samples lead to improved performance, even over the baseline LLaVA-v1.5-7B model.

The second ablation focused on the $\lambda$ parameter, which controls the weight of the vision modeling loss, as outlined in Section \ref{subsec:results_train} and equation \ref{eq:loss}. The optimal performance was obtained with $\lambda=0.1$. As shown in Figure \ref{fig:ablation_lambda}, our approach also outperforms the baseline for all $\lambda \le 0.3$.

Finally, since we are introducing a new capability in the training, we experimented with different learning rates to see if it would lead to improved convergence or better overall performance. Figure \ref{fig:ablation_lr} shows that overall the original learning rate of $2e-5$ achieved the best performance.

\section{Performance of the segmentation maps inherently provided by our model}
\label{appendix:segmentation_analysis}
\begin{figure*}[ht!]
    \centering
    
    \begin{subfigure}[b]{0.3\textwidth}
        \centering
        \includegraphics[width=\textwidth]{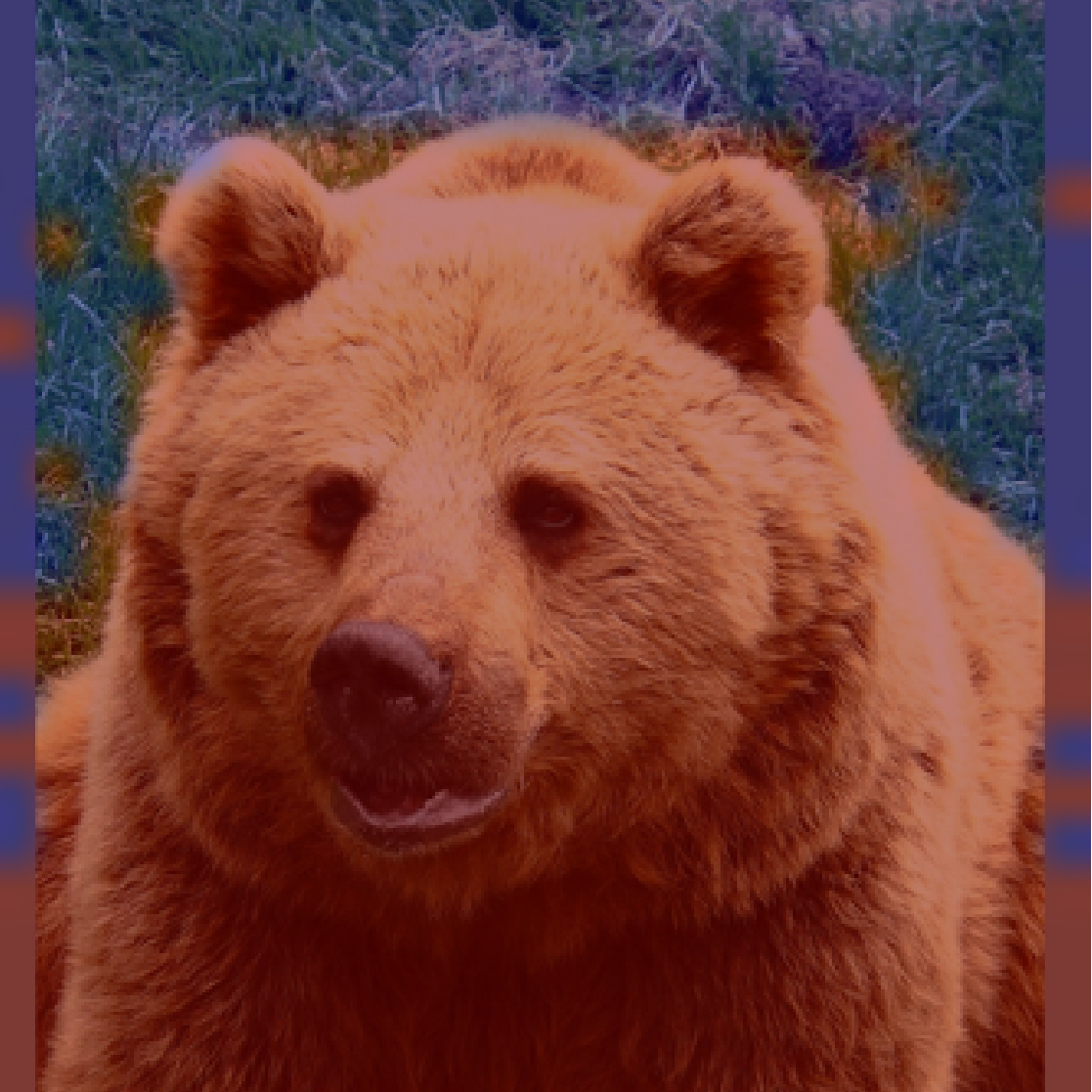}
        \caption{Bear}
        \label{bear}
    \end{subfigure}
    \hfill
    \begin{subfigure}[b]{0.3\textwidth}
        \centering
        \includegraphics[width=\textwidth]{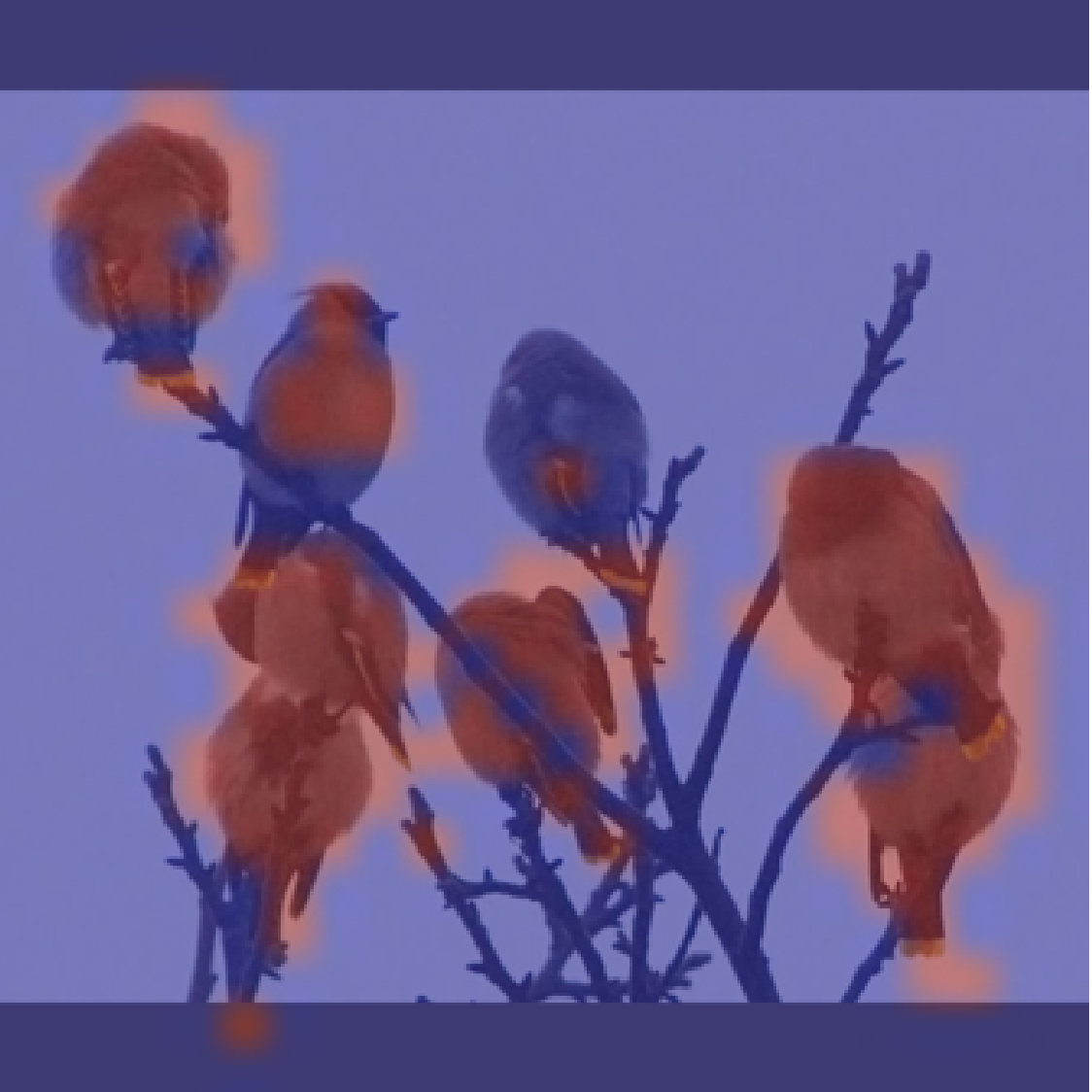}
        \caption{Bird}
    \end{subfigure}
    \hfill
    \begin{subfigure}[b]{0.3\textwidth}
        \centering
        \includegraphics[width=\textwidth]{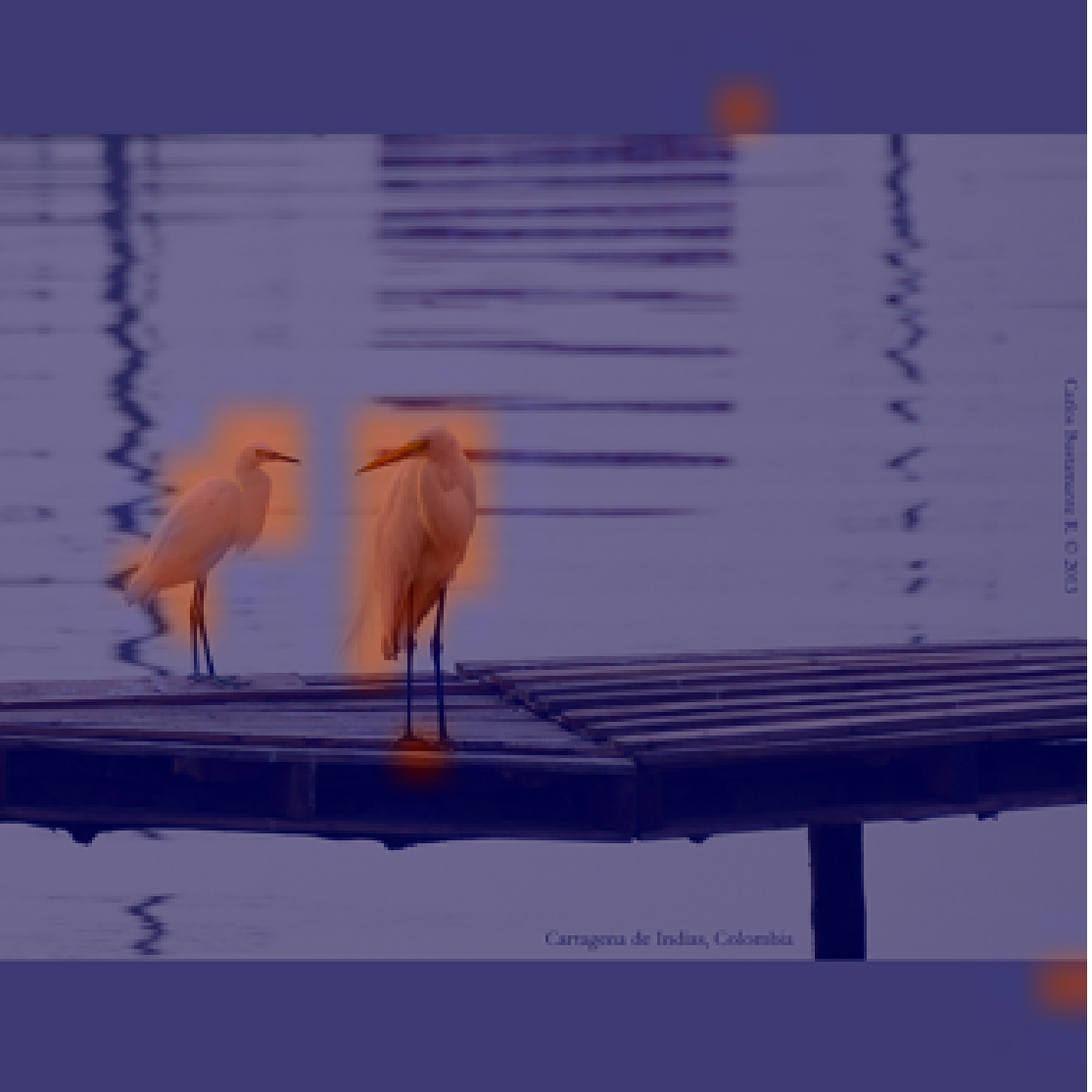}
        \caption{Birds}
    \end{subfigure}

    \vspace{0.5cm} 

    \begin{subfigure}[b]{0.3\textwidth}
        \centering
        \includegraphics[width=\textwidth]{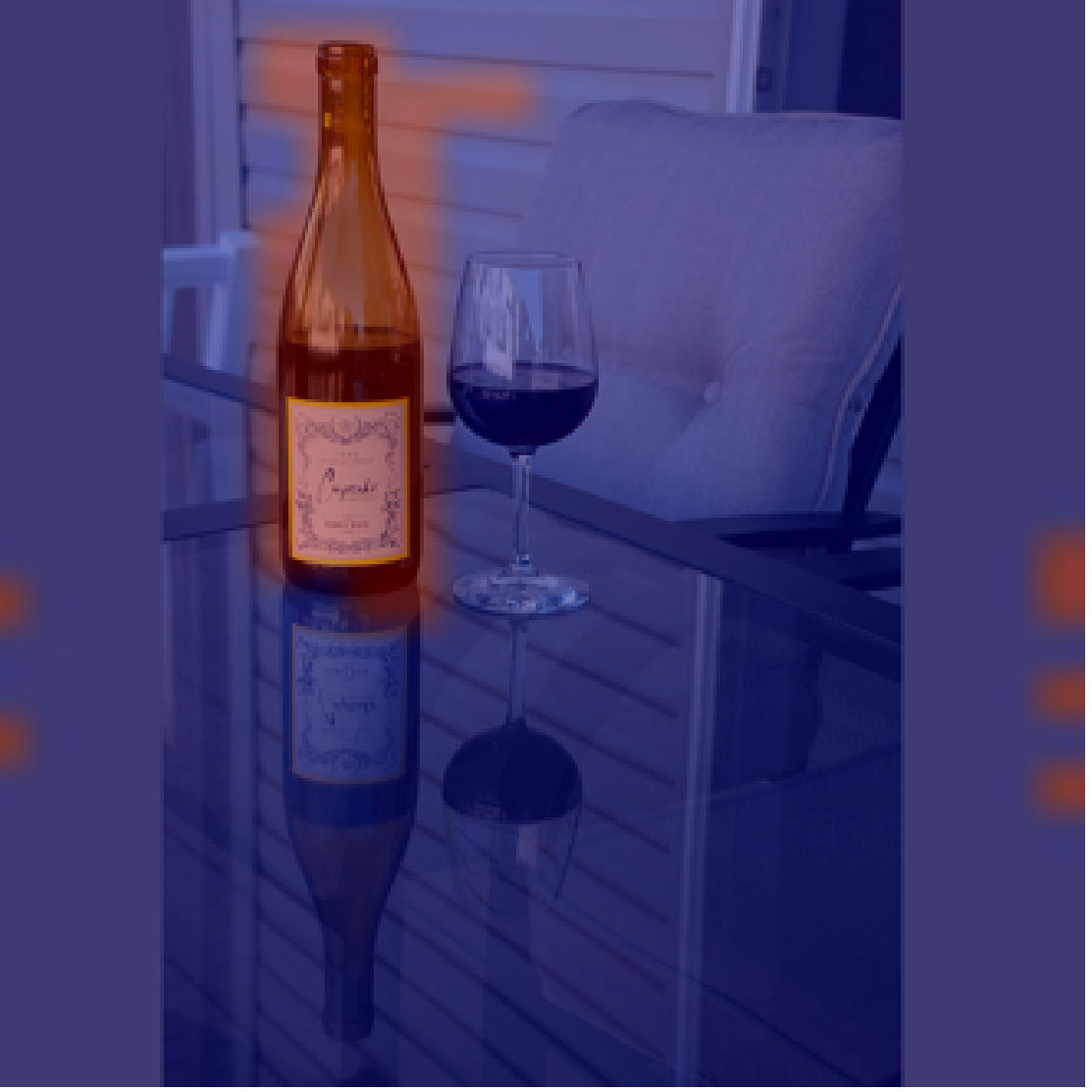}
        \caption{Bott}
    \end{subfigure}
    \hfill
    \begin{subfigure}[b]{0.3\textwidth}
        \centering
        \includegraphics[width=\textwidth]{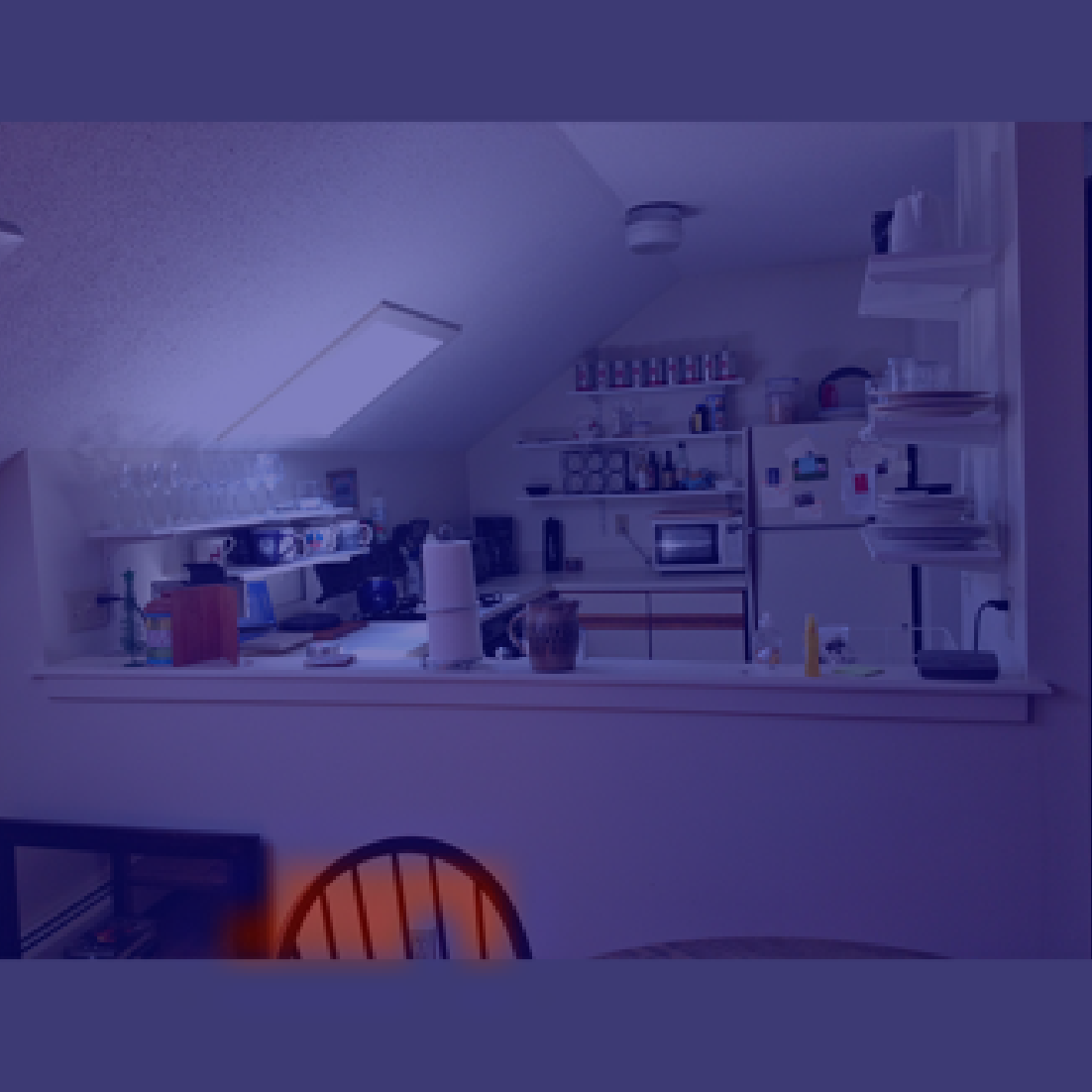}
        \caption{Chair}
    \end{subfigure}
    \hfill
    \begin{subfigure}[b]{0.3\textwidth}
        \centering
        \includegraphics[width=\textwidth]{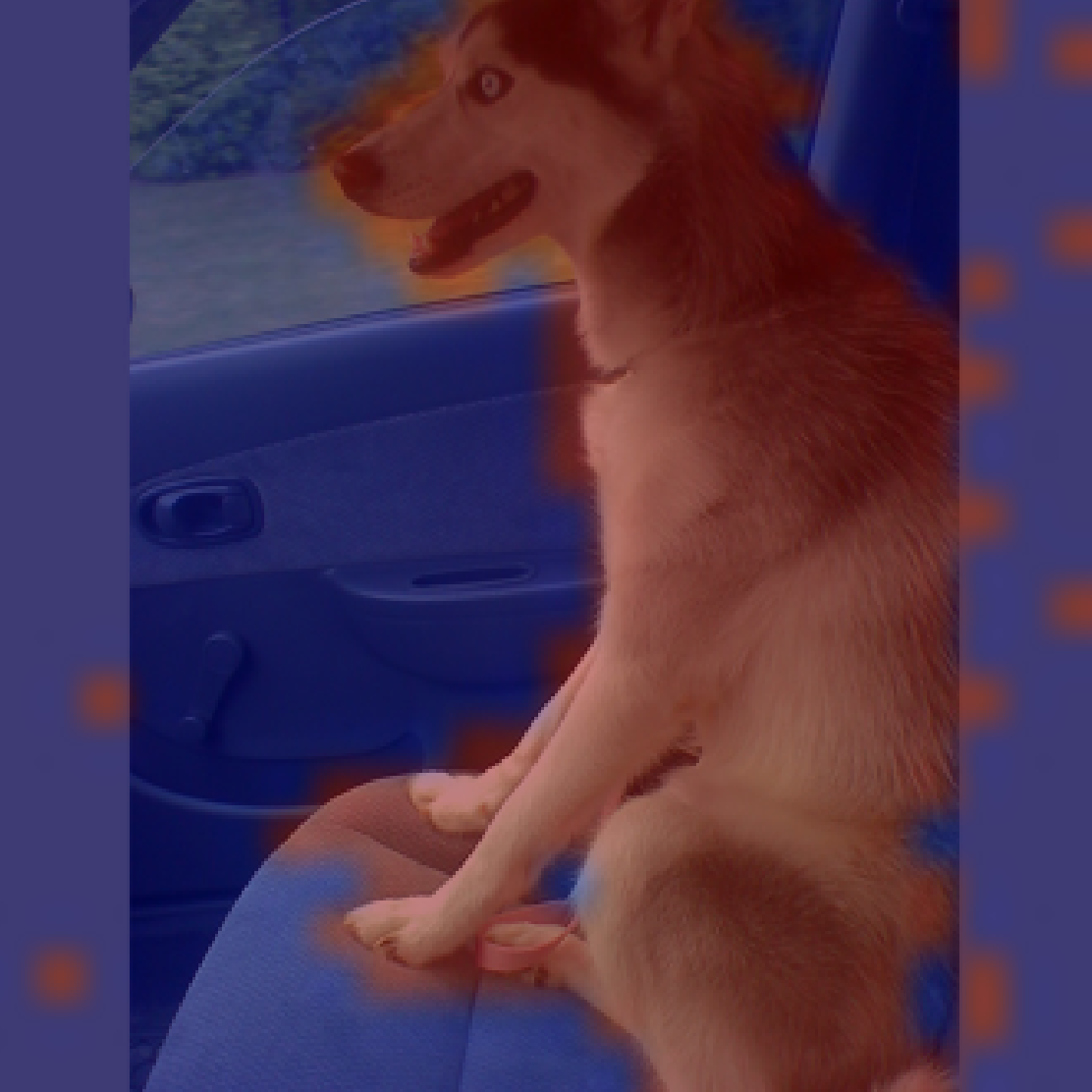}
        \caption{Dog}
    \end{subfigure}

    \vspace{0.5cm} 

    \begin{subfigure}[b]{0.3\textwidth}
        \centering
        \includegraphics[width=\textwidth]{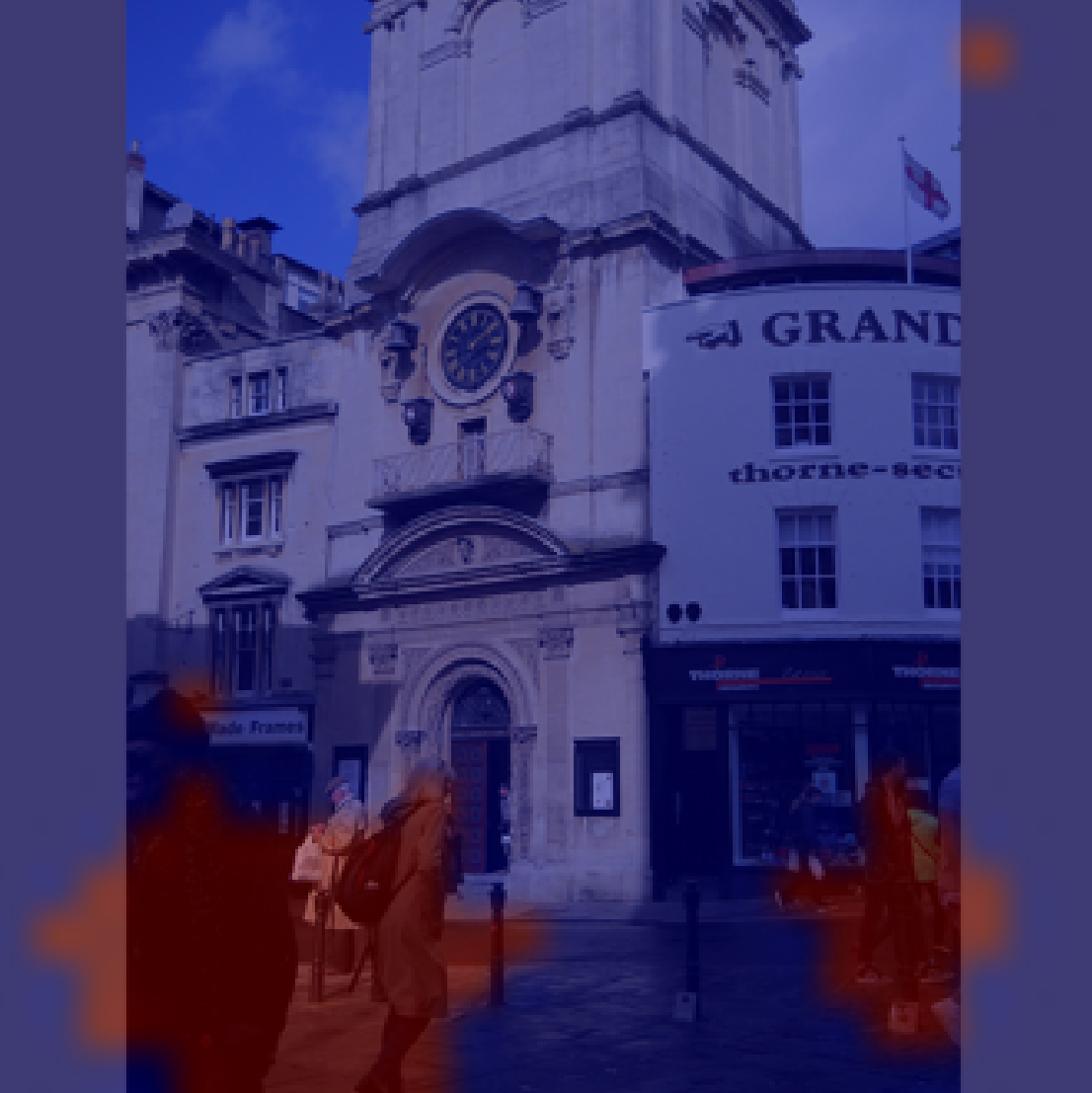}
        \caption{People}
    \end{subfigure}
    \hfill
    \begin{subfigure}[b]{0.3\textwidth}
        \centering
        \includegraphics[width=\textwidth]{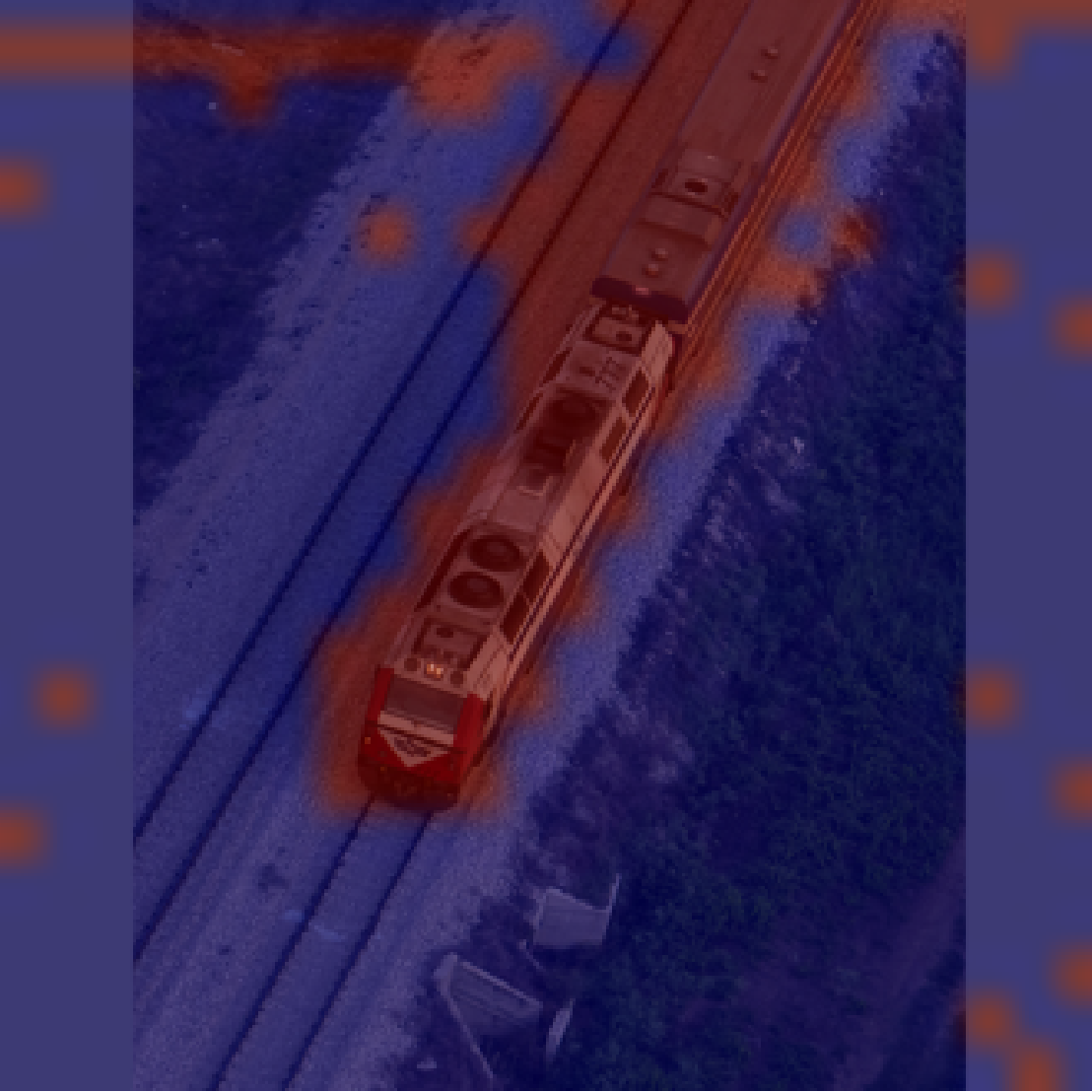}
        \caption{Train}
    \end{subfigure}
    \hfill
    \begin{subfigure}[b]{0.3\textwidth}
        \centering
        \includegraphics[width=\textwidth]{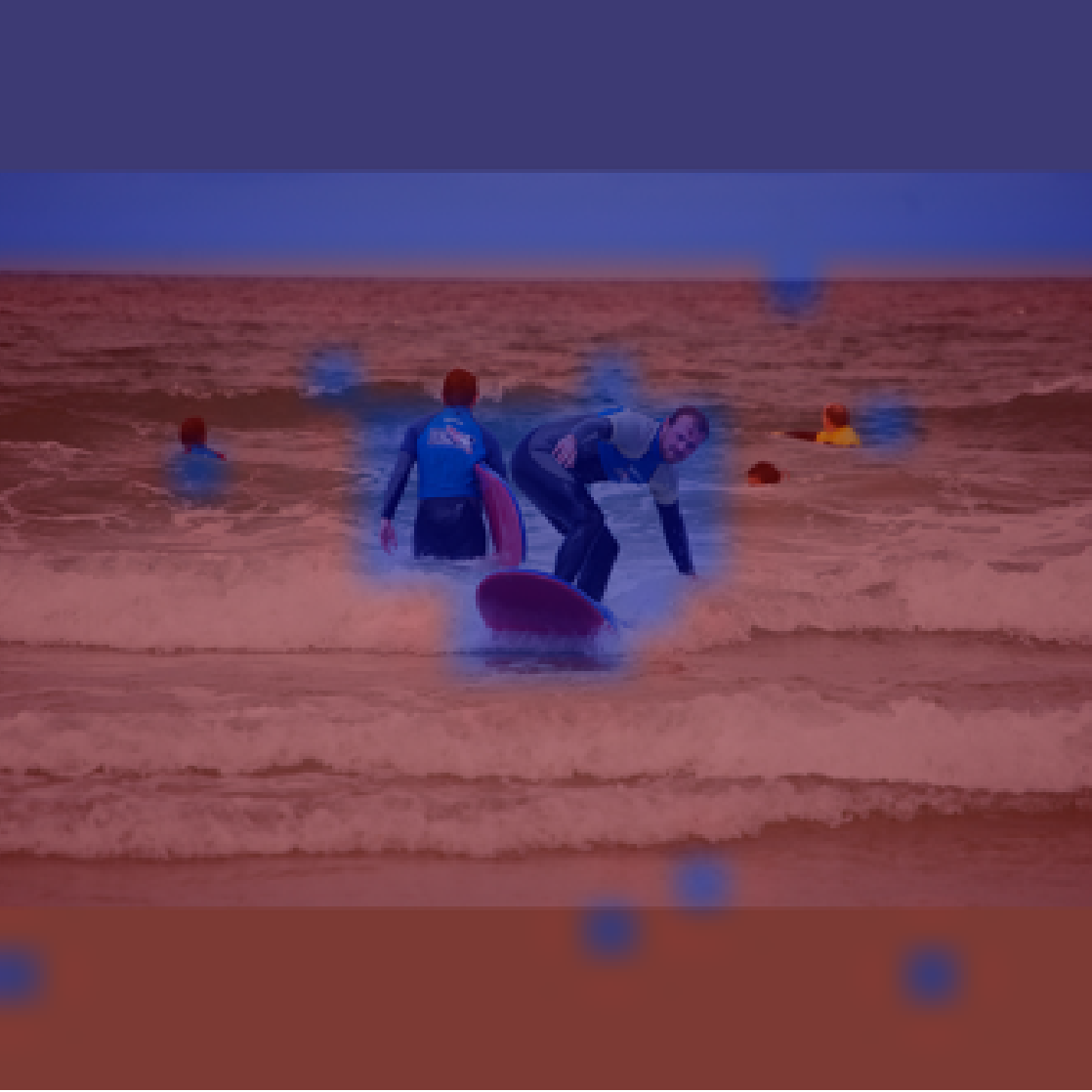}
        \caption{Water}
    \end{subfigure}

    \caption{Segmentations produced inherently by our model. Each figure corresponds to the max-v token specified in caption. Max-v token being the token realizing the maximum for each highlighted patch}
    \label{fig:segmentation_images}
\end{figure*}

To evaluate the segmentation ability of our FiVL model, we evaluated Intersection-Over-Union (IoU) on a subset of 10,000 images from the GQA-val dataset. For each sample, we perform an inference using the baseline LLaVA-7b and our model. From the outputs, we retrieve the visual logits for each visual token, we assigned a text token from the vocabulary corresponding to the maximum logit probability, referred to as the max-v token. By aggregating all image tokens associated with each max-v token, we effectively generated a segmentation mask for each represented text token, like describe in Section 5.1.
Additionally, as ground truth to compare against, we employed Grounded-SAM to produce segmentation maps given each max-v token. Grounded-SAM was implemented using the IDEA-Research/grounding-Dino-Tiny model with thresholds set at 0.2, 0.4, and 0.6, followed by facebook/sam-vit-huge with a threshold of 0.0. The Intersection over Union (IoU) score was computed between the FiVL-generated segmentation masks and the corresponding Grounded-SAM masks to quantitatively assess alignment. To provide a comparative analysis, we also computed IoU scores for the segmentation masks produced by the baseline model.
As detailed in Table \ref{tab:segmentation_iou}, across all thresholds, FiVL generated approximately 7 times fewer max-v tokens per image compared to the baseline model (column  \#tokens/sample), indicating more concise and semantically meaningful segmentation. FiVL also showed significant improvement in average IoU scores (column IoU), increasing approximately three times: from 0.05 to 0.18 at a threshold of 0.2, from 0.06 to 0.21 at 0.4, and from 0.09 to 0.24 at 0.6, showcasing its superior ability to generate precise and coherent segmentation masks. In general, across all thresholds, the baseline generates significantly more max-v tokens per image, resulting in a higher number of samples with segmentation maps found by Grounded-SAM (column \#samples). Finally, the percentage of tokens processed by Grounded-SAM is substantially higher for our model compared to the baseline (column \#processed), indicating that the max-v tokens retrieved by our model were more meaningful than those from the baseline. Figure \ref{fig:segmentation_images} shows the segmentation maps we obtained for the max-v token describing each image. For example for the example \ref{bear}, we computed the argmax of the tokens highlighted in red, and it corresponded to the token "bear" in the vocabulary

\begin{table}[h]
    \centering
    \resizebox{\columnwidth}{!}{%
    \small
    \begin{tabular}{|c|c|c|c|c|c|}
        \hline
         & Thresh & IoU & \# tokens/sample & \# samples & \#processed \\ \hline

        Baseline    & \multirow{2}{*}{0.2}   & 0.05 & 73.3 & 10,000 & 0.89 \\
        Our Model    &       & \textbf{0.18}     & 10.3     & 10,000    & 0.96     \\ \hline
        Baseline & \multirow{2}{*}{0.4} & 0.06 & 73.3 & 10,000 & 0.40 \\
        Our Model     &       & \textbf{0.21}     & 10.3     & 9,983   & 0.65    \\ \hline
        Baseline    & \multirow{2}{*}{0.6} & 0.09 & 73.4 & 9,326 & 0.08 \\ 
        Our Model    &       & \textbf{0.24}    & 10.6    & 8,604    & 0.26    \\ \hline
    \end{tabular} %
    }
    \caption{Performance of the segmentation maps inherently provided by our model}
    \label{tab:segmentation_iou}
\end{table}

\begin{figure}[th]
\begin{tcolorbox}[title=FiVL-Instruct system prompt]
    \small
\texttt{A multimodal instruction-following dataset used for visual instruction tuning and it contains an image and a conversation. The conversation is constructed from a few turns of questions and answers regarding the image.}

\texttt{Given only a question and answer pair: identify short expressions from the answer which could not be generated without the image. }

\texttt{The expression }
\begin{itemize}
    \item \texttt{expresses a visual content from the image.}
    \item \texttt{should be as short as possible.}
    \item \texttt{should not be longer than 4 words}
    \item \texttt{should not include punctuation}
    \item \texttt{should no include reference to the image}
\end{itemize}
\texttt{Unrelated expressions should be separated by the following string: ":::"}

\texttt{Don't add any additional information to the prompt.}

\texttt{For example:}

\texttt{Q: What are the giraffes doing in the image? $<$image$>$}

\texttt{A: The baby giraffe is walking next to the mother giraffe, both moving through the open area of their enclosure}

\texttt{The output should be as following:}

\texttt{baby giraffe:::mother giraffe :::open area of their enclosure}

\texttt{Identify the tokens for the following:}

   \texttt{Q: \{question\} }
   
   \texttt{A: \{answer\}
    }
\end{tcolorbox}
\caption{FiVL-Instruct system prompt used for training datasets}
\label{prompt:instruct-system-prompt}
 \end{figure}

\begin{figure}[th]
\begin{tcolorbox}[title=FiVL system prompts for the evaluation datasets]
\small        
\texttt{A multimodal instruction-following dataset used for visual instruction tuning and it contains an image and a conversation. The conversation is constructed from a few turns of questions and answers regarding the image.}
    
\texttt{Given only a question and answer pair: identify short expressions from the answer or the question which could not be generated without the image.}

\texttt{The expression}
\begin{itemize}
    \item \texttt{should hypothetically express an immediate visual content from image. Thus, yes/no is NOT an expected expression, and some pronouns like "this", "that", "there", and "those" are not expected expressions.}
    \item \texttt{should be as short as possible. }
    \item \texttt{should not be longer than 4 words.}
    \item \texttt{should not include punctuations.}
    \item \texttt{should not include reference to the image, like: "the man wearing a blue suit in the image" it should be: "man wearing blue suit".}
\end{itemize}

\texttt{Unrelated expressions should be separated by the following string: ":::".}

\texttt{If there is no expressions found, your response must be "N/A".}

\texttt{Do not add any additional information to the prompt.}
\\ 

\texttt{\textbf{<EXAMPLES>}}\\ \\

\texttt{Identify the expressions for the following:}\\
\texttt{Q: "\{question\}"} \\
\texttt{A: "\{answer\}"} \\
\end{tcolorbox}
\caption{FiVL-Instruct system prompt used for the evaluation datasets}
\label{prompt:eval_datasets}
\end{figure}

\begin{figure}[th!]
\begin{tcolorbox}[title=FiVL-VQAv2 - Examples]
    \small
    
\texttt{\textbf{Example 1:}}

\texttt{\textbf{Q}: "Considering the limited space and packed fixtures, what suggestions can be provided for organizing the bathroom to optimize its use?"}

\texttt{\textbf{A}: "To optimize the use of the small bathroom with tightly packed fixtures, there are some organization solutions that can be recommended: 
Install wall-mounted storage: By using vertical space on the walls and areas above the toilet, you can create additional storage for toiletries and other bathroom essentials without taking up floor space.
Utilize narrow shelves or over-the-door organizers: Placing narrow shelves or over-the-door organizers on available narrow spaces can help with the organization of toiletries and other small items. 
Use multi-functional items: Opt for items that serve more than one purpose, such as a toilet paper holder with a shelf or a mirror with built-in storage. Opt for compact accessories: Choose smaller or more compact versions of bathroom essentials, like toothbrush holders or soap dishes, to maximize space on countertops and around the sink. Declutter regularly: Frequently assess your bathroom supplies and remove any items that are not in use or have expired to keep the bathroom tidy and make the most of the limited space. By following these suggestions, the bathroom can appear less cluttered, and occupants can make better use of the available space."}

\texttt{\textbf{Key Expressions}: small bathroom:::tightly packed fixtures:::vertical space on the walls:::areas above the toilet:::over-the-door organizers:::narrow spaces:::space on countertops and around the sink}
\\

\textbf{Example 2}:

\texttt{\textbf{Q}: "How  many shadows are on the ground?"}

\texttt{\textbf{A}: "3"}

\texttt{\textbf{Key Expressions}:shadows:::3}
\end{tcolorbox}
\caption{Examples for FiVL-VQAv2}
\label{prompt:ex_vqa}
\end{figure}

\begin{figure}[th!]
\begin{tcolorbox}[title=FiVL-GQA and FiVL-POPE - Examples]
    \small
    
\texttt{\textbf{Example 1:}}

\texttt{\textbf{Q}: "Considering the limited space and packed fixtures, what suggestions can be provided for organizing the bathroom to optimize its use?"}

\texttt{\textbf{A}: "To optimize the use of the small bathroom with tightly packed fixtures, there are some organization solutions that can be recommended: 
Install wall-mounted storage: By using vertical space on the walls and areas above the toilet, you can create additional storage for toiletries and other bathroom essentials without taking up floor space.
Utilize narrow shelves or over-the-door organizers: Placing narrow shelves or over-the-door organizers on available narrow spaces can help with the organization of toiletries and other small items. Use multi-functional items: Opt for items that serve more than one purpose, such as a toilet paper holder with a shelf or a mirror with built-in storage. Opt for compact accessories: Choose smaller or more compact versions of bathroom essentials, like toothbrush holders or soap dishes, to maximize space on countertops and around the sink. Declutter regularly: Frequently assess your bathroom supplies and remove any items that are not in use or have expired to keep the bathroom tidy and make the most of the limited space. By following these suggestions, the bathroom can appear less cluttered, and occupants can make better use of the available space."}

\texttt{\textbf{Key Expressions:} small bathroom:::tightly packed fixtures:::vertical space on the walls:::areas above the toilet:::over-the-door organizers:::narrow spaces:::space on countertops and around the sink}
\\
\texttt{\textbf{Example 2:}}

\texttt{\textbf{Q}: "Is there a snowboard in the image?"}

\texttt{\textbf{A}: "no"}

\texttt{\textbf{Key Expressions:} snowboard}

\end{tcolorbox}
\caption{Examples for GQA and POPE prompts}
\label{prompt:ex_gqa_pope}
\end{figure}


\section{API of the manual evaluation}
\label{appendix:api}
\begin{figure}[h]
    \centering
    \includegraphics[width=\textwidth]{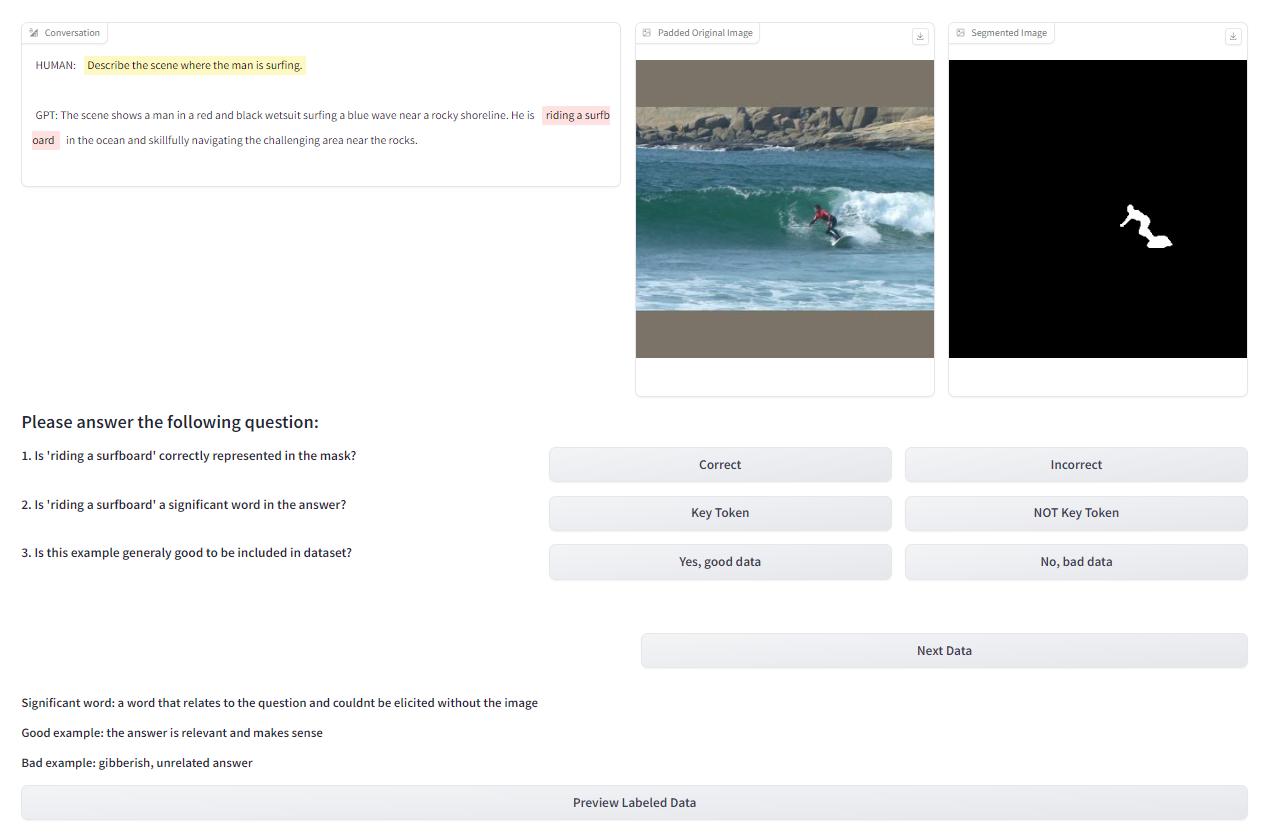}
    \caption{Web user interface for our dataset evaluation}
    \label{fig:api}
\end{figure}

Figure \ref{fig:api} shows the API used for the manual evaluation done on FiVL-Instruct. Given a question, an answer (on the left) and an image with a segmentation mask (on the right), the annotator had to answer the 3 following yes/no questions: is \{ key expression \} correctly represented in the mask? Is \{key expression\} a significant word in the answer? Is this example generally good to be included in the dataset?

\end{document}